%% file: BAM.tex
\documentclass[accepted]{class2024}
\usepackage[american]{babel}
\usepackage{natbib} 
\usepackage{mathtools} 
\usepackage{booktabs} 
\usepackage{tikz} 
\usepackage{amsfonts} 
\usepackage{nicefrac} 
\usepackage{microtype}
\usepackage{xcolor}
\usepackage[nolist]{acronym}
\usepackage{amssymb}
\usepackage{amsthm}
\usepackage{subcaption}
\usepackage{float}

\newcommand{\bs}[1]{\boldsymbol{#1}} 
\newcommand{\bt}[1]{\widetilde{\boldsymbol{#1}}} 
\newcommand{\spaceSPD}[0]{\mathcal{S}^{d\times d}_{\succeq}}
\newcommand{\spaceSPDC}[0]{\mathcal{S}^{d\times d\times C}_{\succeq}}
\newcommand{\Nleft}{\mathopen{}\mathclose\bgroup\left}
\newcommand{\Nright}{\aftergroup\egroup\right}
\newcommand{\citePos}[2]{(#2 \citeauthor{#1}, \citeyear{#1})}
\makeatletter

\newcommand*{\centernot}{
  \mathpalette\@centernot
}
\def\@centernot#1#2{
  \mathrel{
    \rlap{
      \settowidth\dimen@{$\m@th#1{#2}$}
      \kern.5\dimen@
      \settowidth\dimen@{$\m@th#1=$}
      \kern-.5\dimen@
      $\m@th#1\not$
    }
    {#2}
  }
}
\makeatother

\makeatletter 

\newcommand*\sNeg[2][0mu]{\Neginternal{#1}{\snegslash}{#2}}

\newcommand*\Neginternal[3]{\mathpalette\Neg@{{#1}{#2}{#3}}}
\newcommand*\Neg@[2]{\Neg@@{#1}#2}
\newcommand*\Neg@@[4]{
  \mathrel{\ooalign{
    $\m@th#1#4$\cr
    \hidewidth$\m@th#3{#1}\mkern\muexpr#2*2$\hidewidth\cr
  }}
}

\newcommand*\negslash[1]{\m@th#1\not\mathrel{\phantom{=}}}
\newcommand*\snegslash[1]{\rotatebox[origin=c]{60}{$\m@th#1-$}}
\newcommand*\ssnegslash[1]{\rotatebox[origin=c]{60}{$\m@th#1{\dabar@}\mkern-7mu{\dabar@}$}}
\newcommand*\sssnegslash[1]{\rotatebox[origin=c]{60}{$\m@th#1\dabar@$}}
\makeatother

\newcommand{\independent}{\perp\mkern-9.5mu\perp}
\newcommand{\notindependent}{\centernot{\independent}}

\theoremstyle{plain}
\newtheorem{theorem}{Theorem}
\newtheorem{proposition}[theorem]{Proposition}

\theoremstyle{definition}

\theoremstyle{remark}

\input{acronyms.tex}

\makeatletter
\renewcommand\AB@authnote[1]{} 
\renewcommand\AB@affilnote[1]{} 
\makeatother

\title{Graph Structure Inference with BAM: \\Introducing the Bilinear Attention Mechanism}

\author{Philipp~Froehlich \quad Heinz~Koeppl}

\affil{Department of Electrical Engineering and Information Technology\par
Technische Universität Darmstadt, Germany}

\begin{document}
\maketitle

\begin{abstract}
  In statistics and machine learning, detecting dependencies in datasets is a central challenge. We propose a novel neural network model for supervised graph structure learning, i.e., the process of learning a mapping between observational data and their underlying dependence structure. 
  The model is trained with variably shaped and coupled simulated input data and requires only a single forward pass through the trained network for inference. By leveraging structural equation models and employing randomly generated multivariate Chebyshev polynomials for the simulation of training data, our method demonstrates robust generalizability across both linear and various types of non-linear dependencies. We introduce a novel bilinear attention mechanism (BAM) for explicit processing of dependency information, which operates on the level of covariance matrices of transformed data and respects the geometry of the manifold of symmetric positive definite matrices. Empirical evaluation demonstrates the robustness of our method in detecting a wide range of dependencies, excelling in undirected graph estimation and proving competitive in completed partially directed acyclic graph estimation through a novel two-step approach.
\end{abstract}

\section{Introduction}

\label{Introduction}
The inference of causal relationships is central to various scientific fields like biology~\citep{Buehlmann2014annual, Jones2012}, climate science ~\citep{nowack2020causal}, economics~\citep{barfuss2016parsimonious}, and social studies~\citep{Gerstenberg2020}. These relationships are often represented as directed edges within \iac{dag}, a methodology pioneered in agricultural research about a century ago by \citet{Wright1921CorrelationAndCausation} and widely adopted today. One key application is in estimating gene regulatory networks from experimental data  ~\citep{spirtes2000constructing,LauritzenLocal}. Graph structure inference, the process of deriving such graphical representation from observational data, is crucial for gaining insights into high-dynamical systems \citep{Glymour2019}. 

Graph structure inference typically employs unsupervised learning methods to estimate the underlying graph through either score-based approaches, which rank graphs by predefined metrics, or constraint-based approaches that determine edge existence between variable pairs using conditional independence tests \citep{Vowels2022}. 
However, these methods face challenges. Score-based approaches encounter computational burdens due to the superexponential growth of potential graph structures with node count, and the necessity to balance fit and structural sparsity \citep{ke2022learning}. Constraint-based methods usually require a large sample size \citep{Vowels2022}, rely on an elusive optimal threshold hyperparameter\footnote{While a neural network approach also involves hyperparameters, they are primarily involved in the training process, such that they are less sensitive and less uncertain compared to those in constraint-based methods, which directly influence the sparsity of the matrices predicted by the model.}, and \citet{shah2020} proved that the failure of Type I error control in underlying conditional independence tests is unavoidable, which can have significant consequences in downstream analyses.

Supervised causal learning techniques, as presented by \citet{lopez2015randomized,lopez2015towards, Lopez2017, li2020supervised,ke2022learning,Lorch2022,Dai2023}, have recently emerged as an appealing alternative to unsupervised methods. In this rising approach, models are typically trained on simulated matrix-shaped data, with corresponding graph structures serving as ground-truth labels for supervised learning. The paradigm capitalizes on the strengths of deep learning to discern complex patterns in data, thereby enabling accurate graph structure inference. It proves effective even with small datasets and provides the option for fine-tuning when labeled data is accessible.

Supervised causal learning techniques strongly rely on the ability of neural networks to extract dependency information from observational data matrices. Moving beyond traditional methods for predicting dependencies of variables using neural networks, exemplified by the works of \citet{ke2022learning,Lorch2022,Rao2021,kossen2021selfattention,song2019}, which typically embed observational data into an expanded space to implicitly capture dependency information within Euclidean space, we introduce a novel observation-to-dependency framework that inputs a data matrix, processes it through parallel channels, and computes transformed covariance matrices. This process leverages a novel attention mechanism tailored to the geometry of \ac{spd} matrix space, enhancing graph structure inference from observational data. Conditional independencies and the behavior of covariance matrices are intrinsically linked, emphasizing the potential of the \ac{spd} matrix space processing approach to efficiently decode essential information for graph structure inference, obscured in observational data. Additionally, utilizing covariance matrices offers a natural method for deriving $d\times d$ adjacency-shaped matrices from $M\times d$ observational data matrices in a permutation- and shape-invariant manner, thereby avoiding the compression of observational data into embedding vectors. 

Since covariance matrices are symmetric, our method is inherently suited for undirected graph estimation and, with an additional step, can infer edge directions. This is achieved through a two-step strategy: first estimating the graph's skeleton and immoralities, then testing these to generate a completed partially directed acyclic graph (CPDAG) estimate, aligning with the PC-algorithm but with reduced computational complexity and minimized errors. Our method's ability to infer symmetrical information simplifies directional inference, offering a comprehensive solution for both undirected and directed graph estimation from observational data.

Lastly, we employ randomly generated multivariate Chebyshev polynomials as dependency functions within structural equation models for training data generation in the supervised approach, using their ability to approximate well-behaved functions with factorially decreasing coefficients.

\paragraph{Background.}
Our neural network architecture incorporates the self-attention mechanism~\citep{Vaswani2017,Bahdanau2015}, a well-established method in \ac{nlp} and computer vision with state-of-the-art performance~\citep{Khan2022}. \citet{kossen2021selfattention} adapted axial attention~\citep{Ho2019} for non-parametric learning with matrix-shaped input, introducing attention mechanisms both between datapoints and between attributes. This axial attention serves as the foundational mechanism for supervised causal discovery methods, as employed by \cite{ke2022learning,Lorch2022}. 

Incorporating manifold constraints into neural network architectures ensures adherence to geometric constraints, drawing on the analysis of retractions, as described by \citet{absil2008optimization}, within the SPD matrix space.

\begin{figure*}[ht!]
\centering
\includegraphics[width=\textwidth]{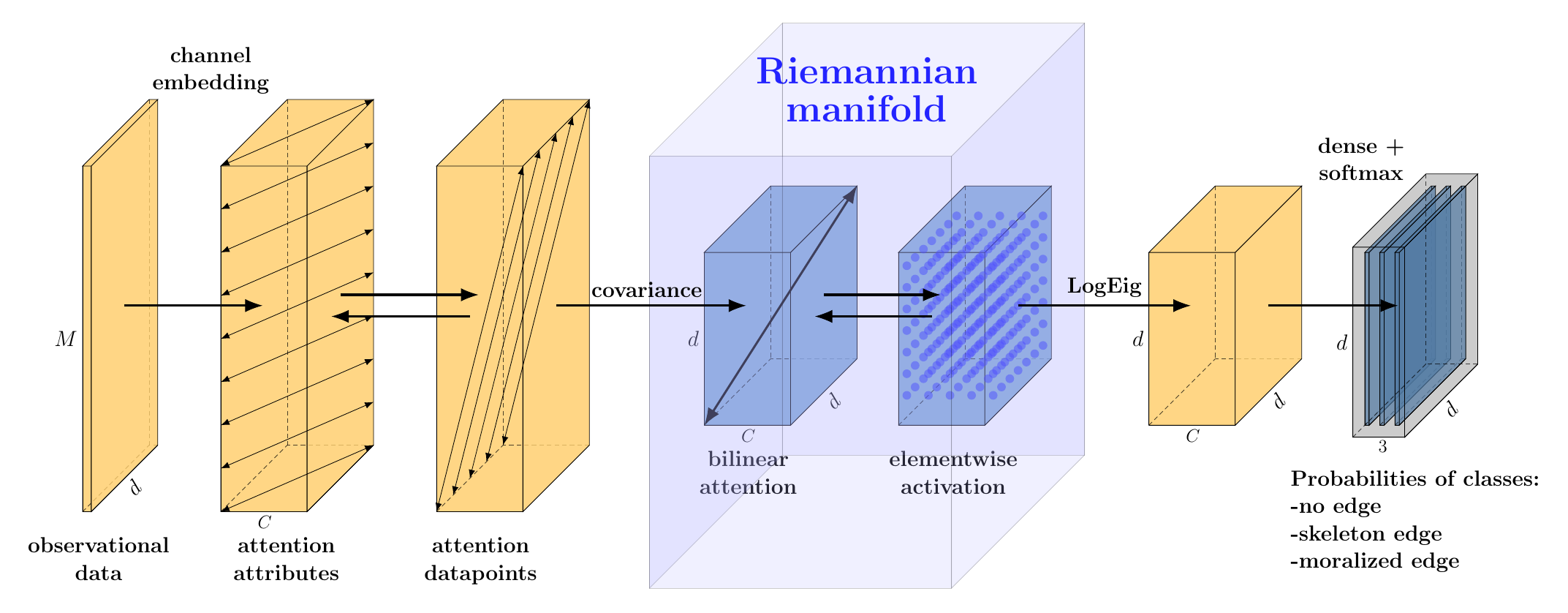}
\caption{Neural network architecture: An input of arbitrarily shape $(M,d)$ is provided, which is then embedded into $C$ channels. Attention between attributes and attention between datapoints are applied alternately. Covariance matrices are calculated, followed by alternating applications of bilinear attention and the custom activation function in the Riemannian manifold of \acf{spd} matrices. The matrices are then transformed into Euclidean space using the $\operatorname{Log-Eig}$ layer. Output probabilities for each pair of variables being in the classes "no edge", "skeleton edge", and "moralized edge" are calculated using dense layers along the channel axis and applying a softmax layer on the channel axis. 
}
\label{fig1}
\end{figure*}

\section{Method}
\subsection{Supervised approach for graph learning}
\paragraph{Problem formulation.}
We consider an underlying \ac{dag} $\mathcal G= (\mathcal V, \mathcal E)$ with nodes $\mathcal V=\{v_1,\dots, v_d\}$ and edges $\mathcal E$. The binary adjacency matrix $\bs{A}\in\{0,1\}^{d\times d}$ denotes the edge presence. For nodes $v \in \mathcal V$, we model the data as a \acf{sem} as $X_v=f_v(X_{\text{pa}_{\mathcal G}(v)},\epsilon_v)$, where $f_v$ is a measurable function of parent nodes and zero-mean error $\epsilon_v$. Each $X_v$ follows a distinct probability distribution determined by the \ac{sem}. Here, $X_{\text{pa}_{\mathcal G}(v)}$ includes random variables in the parent set of a node $v$. The data matrix $\bs{X}\in\mathbb{R}^{M\times d}$ consists of $M$ independent samples from this model.  The aim of the graph inference task is to reconstruct the structure of $\mathcal G$, typically estimating the adjacency matrix $\bs{A}\in\mathbb{R}^{d\times d}$ from data $\bs{X}$. 

\paragraph{Three-class edge classification.}

Our graph structure learning method employs a three-class classification paradigm for the first-step edge inference. We utilize the assumptions of faithfulness and the Markov condition to be able to identify the Markov equivalence class. This enables us to classify pairs of variables distinctly into:

\begin{itemize}
\item \textbf{Skeleton edges}: Representing edges found in the underlying DAG.
\item \textbf{Moralized edges}: Not present in the DAG but emerge due to conditional dependencies among nodes sharing a common child without a connecting edge between the parents.
\item \textbf{No edge}: Signifying variables that remain conditionally independent considering all other variables.
\end{itemize}

Our approach distinguishes between skeleton and moralized edges, modeling them as undirected, which translates to symmetric adjacency matrices. This sets our method apart from some existing algorithms that primarily focus on the moral graph \citep{Friedman2008, cho2014quic, shalom2022pista}.

Given the assumptions stated earlier, the three-class classification problem can be uniquely solved by applying specific independence tests to the data distribution. This can be inferred from the CPDAG's identifiability under these assumptions \citep{spirtes2000causation, peters2017elements}, and the subsequent identifiability of the three-class problem from the CPDAG. In Appendix \ref{Appendix3Class}, we provide an explicit proof, outlining the conditional independence relations that lead to a unique solution.

The targets for prediction can be considered as an extension of the binary adjacency matrix $\bs{A}\in \{0,1\}^{d\times d}$ to a set of one-hot encoded adjacency matrices, denoted as $\widetilde{\bs{A}}\in \{0,1\}^{d\times d\times 3}$. Here, for each $i,j$, the vector $\widetilde{\bs{A}}_{i,j,\cdot}\in\{0,1\}^3$ represents a one-hot encoded classification among the three classes: \emph{skeleton edge}, \emph{moralized edge}, and \emph{no-edge}.

\paragraph{Simulation of training data.}
\begin{figure}[h]
  \includegraphics[width=\columnwidth]{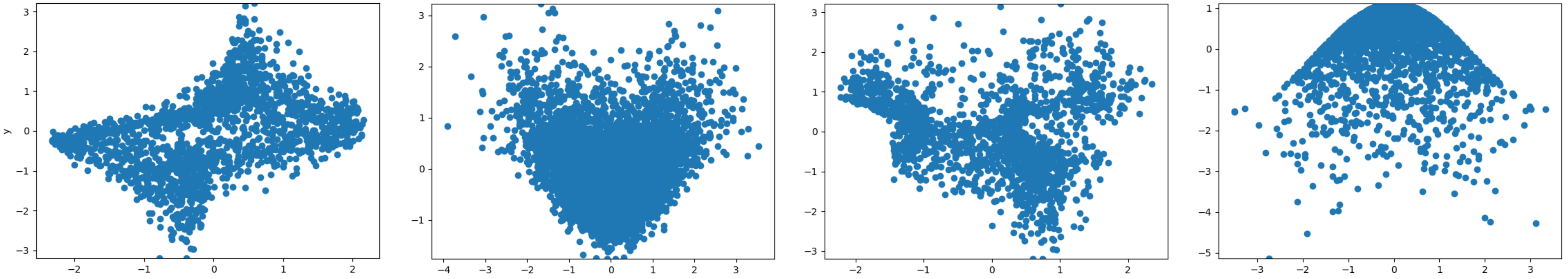}
  \caption{Scatterplots illustrating example non-linear dependencies governed by the structural equation model employed in this study.}
\label{nonlindep}
\end{figure}
To develop a model capable of generalizing across a broad spectrum of functional dependencies and graph structures, we simulate random graphs, denoted $\mathcal{G}_i$, and generate corresponding random data matrix realizations, $\bs{X}_i$ for $i=1,\dots$. These realizations are created from a structural equation model, reflecting the underlying graph structure and incorporating randomly selected dependency functions. This results in an input/label pair $(\bs{X}_i, \mathcal G_i)$ used for supervised learning. To enhance the efficiency of neural network training and prevent overfitting, we employ on-the-fly training, whereby each input/label pair $(\bs{X}_i, \mathcal G_i), i=1,\dots$ is generated just before training and discarded once the neural network weights are updated.

Our approach utilizes random Erdős–Rényi graphs, denoted $ER(d,q)$, where the number of nodes $d$ and the expected degree $q$ are sampled from discrete uniform distributions $d\sim \mathcal U(\{10,\dots,100\})$ and $q\sim U(\{1,\dots,\min(\frac{d}{3},5)\})$, respectively. This configuration allows for denser graphs than those explored in other studies \citep{Dai2023,ke2022learning, yu2019dag}. Our model's versatility enables it to train without restriction to a fixed $(d,q)$ pair, which proves advantageous when the graph density is unknown. For each graph $\mathcal{G}_i$, we generate a data matrix $\bs{X}_i \in\mathbb{R}^{M \times d}$ using \iac{sem}, where the sample size $M$ is drawn from a discrete uniform distribution, $M\sim \mathcal U(\{50,\dots,1000\})$.

We employ random multivariate Chebyshev polynomial functions to generate diverse continuous training data. Chebyshev polynomials effectively approximate real-world functions, showing factorial decay in their coefficients \citep{xiang2020, Trefethen2008}, i.e., for the $n$-th coefficient $c_n$ it holds $\|c_n\|\leq \frac{C}{n!}$ for a constant $C$, making higher-degree terms negligible. In our model, error terms use Gaussian mixture models for a wide range of error distributions \citep{reynolds2009gaussian}. Details about the parameterization of the SEM can be found in Appendix \ref{AppendixSEM}.

\subsection{Shape-agnostic neural network for graph structure learning}
\begin{figure}[h]
    \centering
    \centerline{\includegraphics[width=0.9\columnwidth]{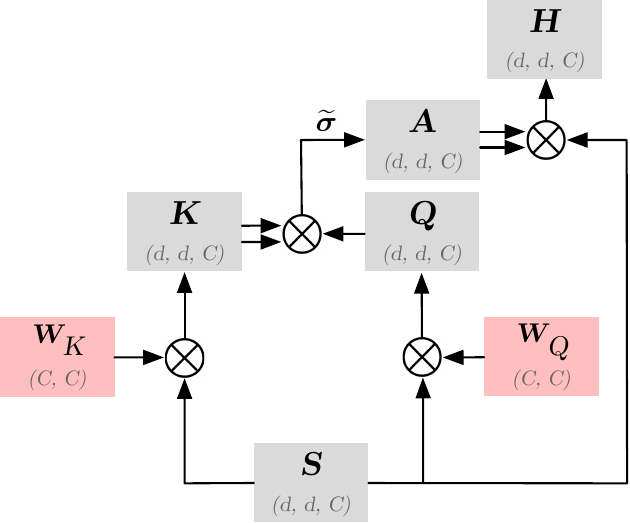}}
\caption{Bilinear self-attention layer. Gray indicates non-trainable tensors, and red trainable weights. Matrix multiplication is performed after necessary transposition to match axis dimensions. The double arrow signifies the use of the matrix as a bilinear operator. $\widetilde{\sigma}$ denotes the custom softmax, defined in \ref{spdmax}.}
    \label{FigAttention}
\end{figure}
For simplicity, we denote each layer's input by $\bs{X}$ or, if in the \ac{spd} manifold, by $\bs{S}$. The output is denoted by $\bs{H}$, so as not to introduce new notation for every layer.
We use the multi-dimensional analogue of matrix multiplication: For a tensor $\bs{A}\in\mathbb{R}^{I \times J \times K}$ and a matrix $\bs{B}\in\mathbb{R}^{K \times L}$ we denote
\[ \bs{A}\bs{B}=\bs{C}\in\mathbb{R}^{I\times J\times L} \quad \text{with}\quad  
\bs{C}_{ijl} = \sum_{k} \bs{A}_{ijk} \bs{B}_{kl}
\]
The network architecture is depicted in Figure \ref{fig1}. 
\paragraph{Channel embedding.}
We perform an embedding of the input $\bs{X}\in\mathbb{R}^{M\times d}$ to obtain a hidden representation with $C$ channels. For this, one axis for $\bs{X}$ is extended to $\bt{X}\in\mathbb{R}^{M\times d \times 1}$ and then trainable weights $\bs{W}_1\in\mathbb{R}^{1\times C}$, $\bs{W}_2\in\mathbb{R}^{C\times C}$ are used to obtain \[\bs{H}=\bt{X}+\text{relu}\left(\bt{X}\bs{W}_1\right)\bs{W}_2\in\mathbb{R}^{M\times d\times C}\] where broadcasting is used for the addition.

\paragraph{Observational data self-attention.}
A description of the observational attention layers is provided in Appendix \ref{AppendixObsAtt}.

\paragraph{Bilinear attention layer.}
After several iterations of data-attention layers, we derive a transformed data matrix $\bs{X}\in\mathbb{R}^{M\times d\times C}$. This matrix is used to compute covariance matrices for each channel, resulting in a tensor of covariance matrices across $C$ channels:
\begin{equation*}
\small
\bs{\Sigma}=\frac{1}{M-1}\left(\left(\bs{X}-\bs{\mu_X}\right)^{T(3,2,1)}\odot\left(\bs{X}-\bs{\mu_X}\right)^{T(3,1,2)}\right)^{T(2,3,1)},
\end{equation*}
where we define for tensors $\bs{A}\in\mathbb{R}^{I\times J\times K}$, $\bs{B}\in\mathbb{R}^{I\times K\times L}$ the $I$-parallel matrix multiplication
\begin{equation*}
\bs{A}\odot\bs{B}:=\Nleft(\bs{A}_{i,\cdot,\cdot} \bs{B}_{i,\cdot,\cdot}\Nright)_{i=1,\dots,I}\in\mathbb{R}^{I\times J\times L},
\end{equation*}
and $\bs{\mu_X}=\frac{1}{M}\bs{1}_M^T\bs{X}\in\mathbb{R}^{1\times d\times C}$ is a tensor of sample means. 

We denote by $\spaceSPD$ the cone of $d\times d$ \ac{spd} matrices, and by $\spaceSPDC=\spaceSPD\times\dots\times\spaceSPD$ we denote the $C$-ary Cartesian power of $\spaceSPD$. It holds $\bs{\Sigma}\in\spaceSPDC$.

Given the nonpositive curvature of the \ac{spd} matrices' Riemannian manifold \citep{bhatia2009}, utilizing SPD matrices in traditional neural networks that operate in Euclidean space poses challenges \citep{pennec2006riemannian, huang2017}. Thus, specialized \ac{spd} networks that respect this manifold's geometry are developed.

Bilinear\footnote{The mapping $\bs{A} \mapsto \bs{A}\bs{\Sigma}\bs{A}^T$ is quadratic, not linear, in $\bs{A}$. However, it's often referred to as 'bilinear' as a specific case of the map $(\bs{A}, \bs{B}) \mapsto \bs{A}\bs{\Sigma}\bs{B}^T$.
} matrix multiplication, analogous to a dense layer in Euclidean space, serves as a primary tool for \ac{spd}-nets \citep{Wang2022}. This is because the mapping $\bs{\Sigma}\rightarrow \bs{A}^T\bs{\Sigma}\bs{A}$, $\spaceSPD\rightarrow  \spaceSPD$ is an endomorphism, i.e., it preserves the space of symmetric, positive semi-definite matrices $\spaceSPD$ of dimension $d$.

In contrast to existing SPD layers, which are typically used for image processing tasks where only a single SPD matrix serves as the covariance descriptor of an image, we do not directly parameterize weights $\mathbf{W} \in \mathbb{R}^{d \times d_{\text{out}}}$ to be applied as $\mathbf{W}^T\mathbf{\Sigma}\mathbf{W}$ to a matrix $\mathbf{\Sigma}\in\mathbb{R}^{d\times d}$. Instead, we parameterize weights $\bs{W}\in\mathbb{R}^{C\times C}$ to act as linear combinations $\bs{S}\bs{W}$ on a set of covariance matrices $\bs{S}\in\spaceSPDC$, leveraging an attention mechanism to create inner weights adaptable to different matrix dimensions $d$ for each SPD matrix in $\bs{S}$.

Our bilinear attention mechanism preserves the positive definiteness of matrices, requiring only a single matrix logarithm computation for the transition to the associated Euclidean space. In this space, the softmax operation over the three classes becomes applicable, enabling the computation of predictions as output. The architecture, illustrated in Figure \ref{FigAttention}, imposes non-negativity constraints on weights $\bs{W}_K\in\mathbb{R}^{C\times C}_+$, $\bs{W}_Q\in\mathbb{R}^{C\times C}_+$ such that $\bs{S}\bs{W}_K, \bs{S}\bs{W}_Q\in \spaceSPDC$, leveraging cone properties. We suggest initializing positive weight matrices $\bs{W}^+$ with samples from $U\Nleft(0, \frac{2}{n_{\text{in}}}\Nright)$, where $n_{\text{in}}$ is the number of input neurons.  In expectation, the diagonal entries in each matrix within $\bs{S}\bs{W}^+$ should stay within a comparable range, while off-diagonal entries are expected to be drawn to zero due to the symmetric distribution of positive and negative values. Consequently, keys and queries have essentially initiated as perturbed identity mappings. 

For input $\bs{S}\in \spaceSPDC$, we obtain keys $\bs{K}=\bs{S}\bs{W}_K\in \spaceSPDC$ and queries $\bs{Q}=\bs{S}\bs{W}_Q\in \spaceSPDC$, which are combined in a bilinear fashion, parallel over the $C$ channels by calculating 
\begin{equation*}
\bs{K}\otimes\bs{Q}:=\left(\bs{K}_{\cdot, \cdot, c} \bs{Q}_{\cdot, \cdot, c} \bs{K}_{\cdot, \cdot, c}\right)_{c=1,\dots, C}\in \spaceSPDC
\end{equation*}
for each channel $c$, where indexing corresponds to the channel axis, results in a tensor of $C$ \ac{spd} matrices. 

To preserve positive definiteness, which is in general compromised by the non-symmetric output of standard softmax application on matrix rows, we propose a custom softmax function. We first define a diagonal scaling matrix as
\[\bs{\Lambda}(\bs{S}):=\operatorname{diag}\Nleft(\frac{1}{{\exp[\bs{S}]\bs{1}_d}}\Nright),\] 
where $\exp[\cdot]$ denotes the elementwise application of the exponential function, $\bs{1}_d$ is a vector of length $d$ with all entries being $1$, and $\operatorname{diag}$ transforms a vector of length $d$ into a $d\times d$ diagonal matrix. The quotient is also taken elementwise.
With this, we propose the custom softmax function $\bt{\sigma}$ as follows:
\begin{gather}
    \label{spdmax}
    \bt{\sigma}: \spaceSPD \rightarrow  \spaceSPD \quad \quad
    \bt{\sigma}(\bs{S}):= 
    \sqrt{\bs{\Lambda}(\bs{S})}\exp[\bs{S}]\sqrt{\bs{\Lambda}(\bs{S})},
\end{gather} 
where the square root is elementwise.

Note that the elementwise application of the exponential function preserves positive definiteness. This is because \ac{spd} matrices are closed under addition and the Hadamard product \citep{bhatia2009}, and it holds that $\exp[\bs{S}]=\sum_{n=0}^\infty \frac{1}{n!}[\bs{S}]^n$, with elementwise exponentiation $[\cdot]^n$. 
Similar to the standard softmax applied over the rows, using this softmax for a $\spaceSPD$ matrix returns positive values that altogether sum to $d$. However, in contrast to the standard softmax, the rows do not sum up to $1$. We demonstrate that our modified softmax function $\bt{\sigma}$ additionally regularizes the eigenvalues:

\begin{theorem}
\label{teig}
For any $\bs{S}\in\spaceSPD$, the largest eigenvalue of $\bt{\sigma}(\bs{S})$ is $1$.
\begin{proof}
Let $\bt{S}:=\exp[\bs{S}]$. By similarity transformation, the eigenvalues of $\bt{\sigma}(\bs{S})$ are equal to the eigenvalues of $\bt{S}\bs{\Lambda}(\bs{S})$. It holds that
$\bt{S}\bs{\Lambda}(\bs{S})\bt{S}\bs{1}=\bt{S}\bs{1}$,
which demonstrates that $\bt{S}\bs{1}$ is the Perron eigenvector corresponding to the eigenvalue $1$. The assertion now follows from the Perron-Frobenius theorem.
\end{proof}
\end{theorem}

\begin{proposition}
\label{prop1}
The custom softmax $\bt{\sigma}$ is invariant to additive shifting, i.e., $\bt{\sigma}(\bs{S}+\alpha)=\bt{\sigma}(\bs{S})$ for each $\alpha\in\mathbb{R}$.
\end{proposition}
Proposition \ref{prop1} shows that $\bt{\sigma}$, unlike standard softmax, does not need a scaling constant, and it can easily manage exploding $\exp$ values via maximum-value scaling.

We obtain the attention matrix $\bs{A}\in \spaceSPDC$ by applying $\bt{\sigma}$ channelwise:  \begin{equation*}
    \bs{A}:=\left(\bt{\sigma}\Nleft((\bs{K}\otimes\bs{Q}\Nright)_{\cdot,\cdot,c})\right)_{c=1,\dots, C}\in \spaceSPDC
\end{equation*}
Finally, the output of the bilinear layer is computed for each channel separately as \[\bs{H}=\bs{A}\otimes\bs{S}\in \spaceSPDC.\] 
\paragraph{\ac{spd} activation function.} We use polynomials of degree $3$ with learnable weights $w_k\geq 0$, $\sum_{k}w_k\leq 1$ as activation function in the \ac{spd} space. For a discussion and a motivation, see Appendix \ref{AppendixActivation}.

\paragraph{Log-Eig layer and output softmax.}
Data representation transitions from the SPD space to Euclidean space through the Log-Eig layer, as \citet{huang2017} proposed. This transformation, given an input matrix $\bs{S}=\bs{U}\bs{D}\bs{U}^T$ via eigendecomposition can be expressed as:
\begin{equation*}
l:\spaceSPD\rightarrow \mathbb{R}^{d\times d},\quad l(\bs{S}):=\bs{\log}\Nleft(\bs{S}\Nright):=\bs{U}\bs{\log}\Nleft(\bs{D}\Nright)\bs{U}^T
\end{equation*}
The final layer, equipped with a softmax activation function, consists of $3$ output units for generating the probabilities.

\paragraph{Interpretation.}
A detailed discussion on the necessity, attention scores, and keys and queries of our novel attention mechanism is available in Appendix \ref{AppendixIntuition}. Unlike traditional attention where the $(i,j)$-th score in the attention matrix $\bs{A}$ indicates the influence of element $i$ on $j$, our bilinear attention mechanism reveals interdependences. For an output pair $(i,j)$, its associated output value is determined not merely by a direct scalar relationship but by the bilinear form $\sum_{k,l} A_{i,k}S_{k,l}A_{l,j}$. 
Consequently, the influence on the $(i,j)$-th entry in the output is based on ${\bs{A}_{i,\cdot} \cup \bs{A}_{\cdot, j}}$ rather than just $A_{i,j}$, leading to attention scores forming a cross shape within the matrix $\bs{A}$. This approach allows columns of the matrix $\mathbf{S}$ to attend to each other, highlighting a more complex interaction pattern.

\paragraph{Implementation details.}
Additional details regarding residual connections, normalization techniques, and the use of multiple heads in attention layers are elaborated in Appendix \ref{AppendixResidual} and \ref{AppendixHeads}.

\paragraph{CPDAG estimation from the graph skeleton and the set of moralized edges.}
To derive the CPDAG from the graph skeleton and identified immoralities, we train a second neural network to infer v-structures from two parent nodes together with potential common child nodes that have edges to both parents, as well as other neighbor nodes related to these parents. We iterate over all inferred immoralities, treating the two nodes involved in each immorality as parent nodes. 
By applying distinct layers to the data corresponding to the parent nodes, potential common children, and neighbors, we can break the symmetry among nodes and enable role-specific learning, thus facilitating edge directionality inference 
The resulting CPDAG is further refined using Meek rules \citep{Meek1995}. Details about the CPDAG estimation step can be found in Appendix \ref{AppendixCPDAG}.

\section{Related Work}
Causal discovery largely relies on unsupervised learning methods, namely constraint-based and score-based approaches \citep{Vowels2022}. The former infer conditional independencies \citep{spirtes2000causation, hyttinen2013, drton2017structure}, while the latter optimize a score function under acyclicity constraints \citep{chickering2002optimal,goudet2018learning}. Continuous optimization techniques within score-based methods have gained attention, utilizing various strategies \citep{zheng2018,yu2019dag,Brouillard2020}. 

Some studies, inspired by the protein contact prediction task that utilizes \ac{msa}, have aimed to estimate relations from \ac{msa} data matrices. For instance, \citet{Rao2021} employed axial attention \citep{Ho2019} on \ac{msa} matrices. Similarly, \citet{Li2019} estimated precision matrices of \acp{msa} and applied a convolutional neural network to predict specific protein contacts. While the latter method shares our intuition of operating on the inverse covariance matrix, our approach significantly differs by transforming data by axial attention prior to covariance matrix calculation and implementing tailored layers for \ac{spd} matrix processing. Moreover, because of the known fixed data dimension of $20$ naturally-occurring residue types, \ac{msa} models are specifically designed for the unique conditions of the MSA prediction task.

Our work is most closely aligned with the studies by \citet{ke2022learning} and \citet{Lorch2022}, which similarly leverage attention mechanisms between samples and attributes to estimate adjacency matrices. However, the methodologies differ in their strategies for deriving an adjacency matrix from observational data representations. \citet{ke2022learning} employ an autoregressive transformer approach, whereas \citet{Lorch2022} utilize the dot product of embedding vectors derived from max-pooling across the sample axis of the observational data matrices. We demonstrate that the derivation of output adjacency matrices can be naturally and efficiently achieved through covariance matrices, which allows for the direct processing of dependence information and enhances learning efficiency through the application of geometric learning on the SPD manifold.

Also, while the model of \citet{ke2022learning} trains different models for varying data dimension, our model can be trained across several different shapes of sample numbers and attribute numbers, thus eliminating the need for re-training on different datasets. Although the model developed by \citet{Lorch2022} is capable of evaluation on data with varying dimensions, their implementation does not support practical training across variable sample sizes due to memory allocation constraints, and it is restricted to handling a limited range of dimensions for training simultaneously.

Additionally, our model addresses identifiability challenges by first estimating an undirected graph, then proceeding to CPDAG estimation employing Meek's rules. This methodology ensures that edge directionality is deduced only when it can be reliably determined, indicated by the identification of v-structures. In contrast, the networks developed by \citet{ke2022learning} and \citet{Lorch2022} deduce a DAG utilizing both observational and interventional data, dependent on the availability of interventional data or making random guesses for edges that cannot be directly inferred.

While \citet{li2020supervised} also used permutation-invariant models for causal inference, their method directly computes correlation matrices, which may limit its efficacy when covariance matrices are not sufficient. They apply layers of the form $\bs{Y}=\omega_1\bs{S}+\omega_2\bs{1}\bs{1}^T\bs{S}+\omega_3\bs{S}\bs{1}\bs{1}^T+\omega_4\bs{1}\bs{1}^T\bs{S}\bs{1}\bs{1}^T+b$, with scalar weights $\omega_i, i=1,\dots, 4$, and bias $b$. Despite preserving invariances, the limited free parameters could restrict its representational power. 

\citet{lopez2015towards,lopez2015randomized} proposed a supervised learning framework that leverages kernel mean embeddings. \citet{Ma2022} address identifiability with independence tests and cascade classifiers for supervised skeleton learning, trained on vicinal graphs specific to observational data. Our method, however, adopts a more general approach, training a network that does not require re-training for new evaluation samples. \citet{Dai2023}'s work on immoralities aligns with our second CPDAG estimation step.

\section{Experiments}\label{secExperiment}
\begin{figure*}[ht!]
\centering
\includegraphics[width=\textwidth]{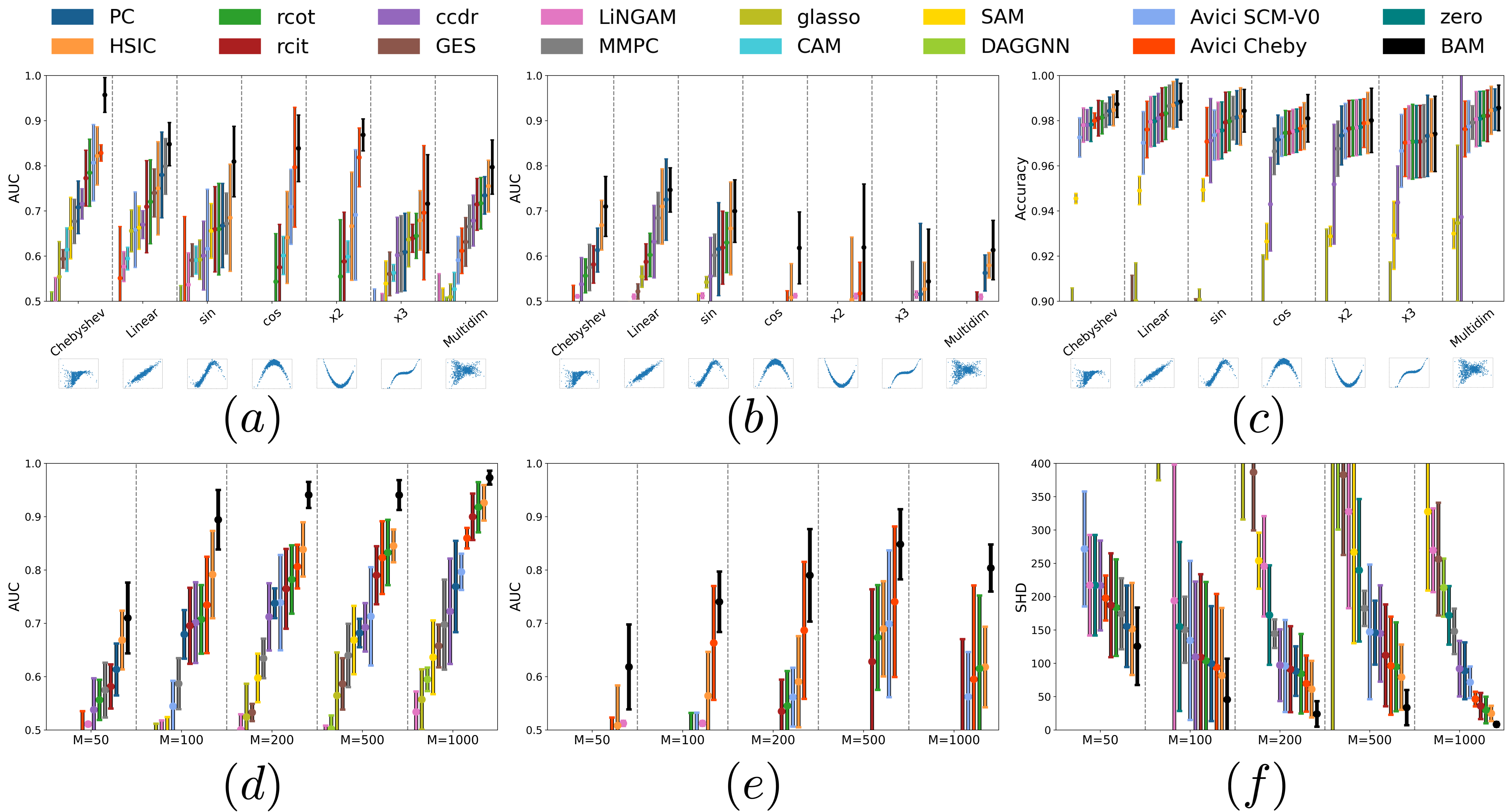}
\caption{Undirected graph estimation results, arranged in each case from worst (left) to best (right). AUC values. (a) and (b) AUC values for different dependencies, with (a) $d=50$, $M=200$ and (b) $d=100$, $M=50$. (c) shows accuracy values for the same dependencies at $d=100$, $M=50$. (d) and (e) present AUC values for different sample sizes with Chebyshev (d) and cosine (e) dependencies at $d=100$. (f) displays structural Hamming distance for varying sample sizes in a high-dimensional setting ($d=100$) for Chebyshev dependency.}
\label{resUnd}
\end{figure*}

\begin{figure*}[ht!]
\centering
\includegraphics[width=\textwidth]{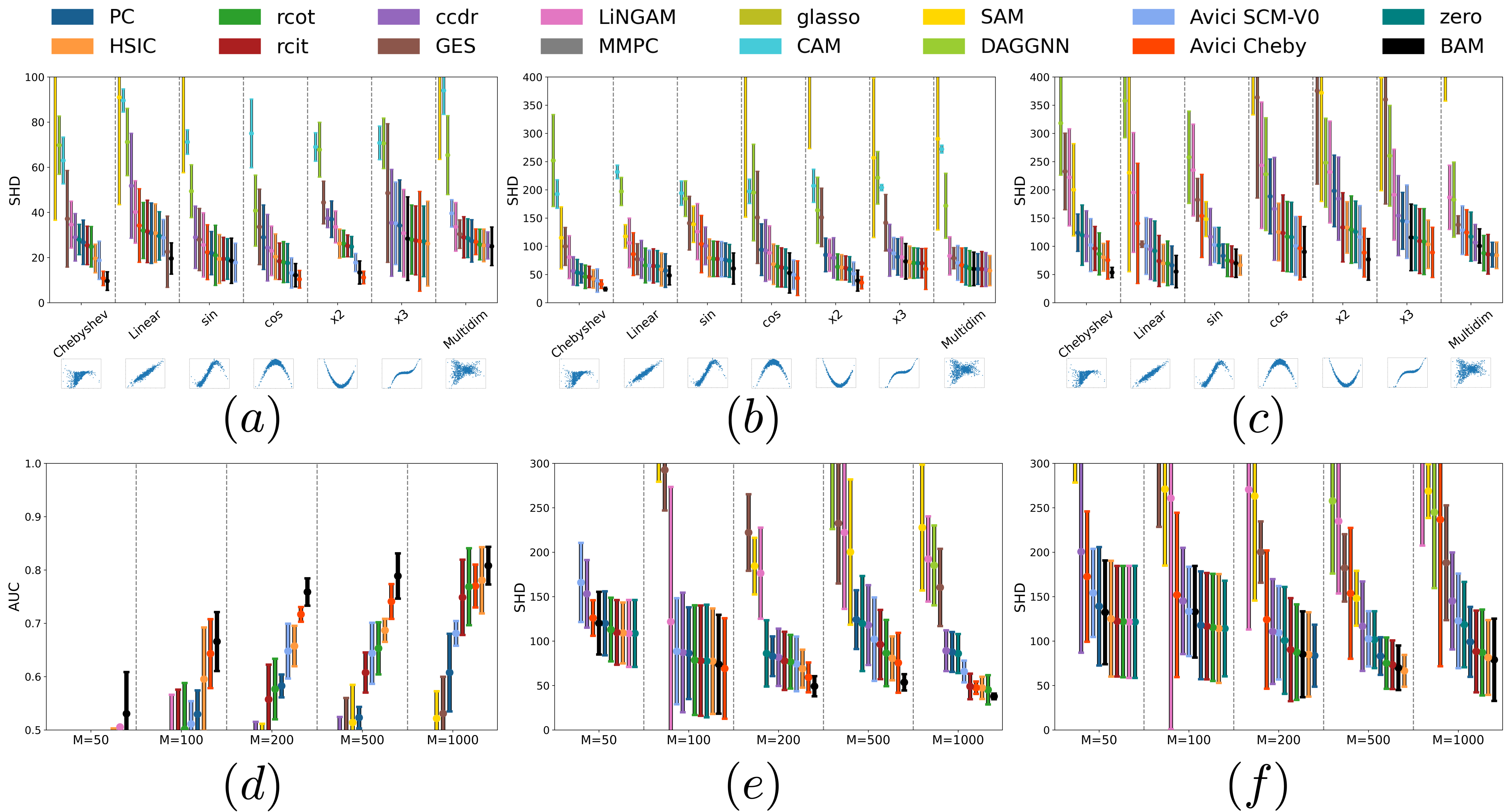}
\caption{CPDAG estimation results ordered from worst (left) to best (right). (a)-(c) SHD for various dependencies at (a) $d=20$, $M=200$, (b) $d=50$, $M=200$, and (c) $d=100$, $M=500$. (d) AUC for Chebyshev dependency across $M=50, 100, 200, 500, 1000$ at $d=100$. (e)+(f) SHD at $d=100$ for Chebyshev and sine dependencies, respectively, over the same $M$ values.}
\label{resCPDAG}
\end{figure*}

\paragraph{Baseline algorithms.} We evaluate BAM's performance against other algorithms: PC \citep{spirtes2000causation}, PC-HSIC \citep{zhang2012kernel}, rcot and rcit \citep{strobl2019approximate}, ccdr \citep{aragam2015concave}, GES \citep{chickering2002optimal}, GIES \citep{hauser2012characterization}, LiNGAM \citep{shimizu2006linear}, MMPC \citep{tsamardinos2006max}, CAM \citep{buhlmann2014cam}, SAM \citep{Kalainathan2022}, all  implemented in the causal discovery toolbox \citep{kalainathan2020causal}, glasso \citep{Friedman2008}, as implemented by \citet{scikit-learn}, along with DAG-GNN \citep{yu2019dag}. For a supervised causal discovery baseline, we use the AVICI model from \citet{Lorch2022}, where we use a pre-trained version (Avici SCM-v0), as well as a version trained from scratch on the same Chebyshev data as our method. For this, we used default hyperparameters, particularly, we used data dimensions $d\in \{2, 5, 10, 20, 30, 40, 60, 80, 100\}$. For the number of samples $M$, which can only be chosen as a single value due to memory allocation constraints, we selected $M$ = 150, as larger sizes exceeded our server’s graphical memory limits (approximately 80 GB).  Default hyperparameters from \citet{kalainathan2020causal} are used. For glasso, cross-validation for the sparsity parameter is performed as proposed by \citep{scikit-learn}. Glasso and MMPC are excluded from CPDAG estimation as they focus on undirected graphs. For algorithms which estimate directed edges, the skeleton is computed for undirected prediction. Addtionally, we compare SHD and accuracy results against a zero graph without any edges as naive baseline.

\paragraph{Performance indicators.}
We employ the Area Under the Precision-Recall Curve (AUC) for evaluating both undirected and CPDAG graph estimations, recognizing its appropriateness for imbalanced binary classification tasks such as sparse graph detection \citep{he2009Learning}. For CPDAG estimations, we further utilize the Structural Hamming Distance (SHD), a benchmark metric in structure learning \citep{yu2019dag, ke2022learning}. For the undirected graph estimation task, SHD is equivalent to accuracy, defined as the percentage of correctly inferred edges, which we also consider in this study.

\paragraph{Results.} Figure~\ref{resUnd} showcases BAM's efficacy in undirected graph prediction. Trained on synthetic Chebyshev polynomial data, BAM was evaluated across various dependency relations, as shown in Figure~\ref{resUnd} (a), (b), (c). It consistently outperforms other methods in both moderate and high dimensions. Remarkably, BAM excels in capturing intricate non-monotonic dependencies such as cosine and $x^2$ dependencies, illustrated in Figure~\ref{resUnd} (a), (b), (e). When applied to mainly monotonic dependencies like sine in Figure~\ref{resUnd} (c), BAM demonstrates superior performance across multiple sample sizes, affirming its generalizability.

To contrast with Avici's distinct model architecture, plots (d) and (f) present the performance for varying sample sizes $M$ on Chebyshev dependency data, utilized in training both BAM and the re-trained Avici version. These results demonstrate BAM's superior performance over Avici on the training set.

Figure~\ref{resCPDAG} displays the results for CPDAG estimation tasks. In high-dimensional scenarios ($d=100$, $M=50$), no algorithm surpassed the baseline of a zero-graph (a graph devoid of edges) in SHD, as shown in Appendix \ref{AppendixDirectedResults}. Therefore, our SHD analysis focuses on the low-dimensional context shown in (a) and the moderate-dimensional context in (b) and in in (c) another regime where $d<M$, specifically $d=100$ and $M=500$, across various dependencies. Across these scenarios, the two-step method of our algorithm remains competitive, though its advantage is less pronounced than in the task of undirected graph estimation. Panels (d) and (e) depict the AUC and SHD across varying $M$ values for the training datasets of BAM and the re-trained Avici, illustrating BAM's competitive performance in learning CPDAG-structure from data. Specifically, panel (e) presents SHD values for sine dependency across different $M$ values for $d=100$, further demonstrating the robustness of BAM in non-linear settings.

\begin{table}[ht!]
  \caption{Ablation study results featuring loss values for evaluation. Data generated under Chebyshev dependencies. "-" indicates the removal of a corresponding layer. $\Delta \text{Param}$ quantifies the difference in the number of parameters to the full model.}
  \label{tabAblation}
  \centering
  \begin{tabular}{lll}
    \toprule
    model     & loss \textdownarrow &$\Delta \text{Param}$\\
    \midrule \\
    FULL  & $0.173\pm 0.007$    \\
    $-$ bilinear  &$0.202\pm 0.005$ &80 K \\
    $-$ bilinear $-$ LogEig &$0.271\pm 0.012$ &100 K \\
    $-$ obs. att.& $0.189\pm 0.006$ & 120 K \\
    $-$ obs. att. $-$ Dense & $0.205\pm 0.006$ & 160 K \\
    \bottomrule
  \end{tabular}
\end{table}

\paragraph{Ablation Study.} Table \ref{tabAblation} presents the results of our ablation study, highlighting the critical role of the bilinear layer. Its removal leads to a significant increase in loss metrics. The omission of the LogEig layer results in a pronounced deterioration of the loss, showing that direct predictions from the SPD-space are leading to problems. Interestingly, bilinear data processing alone, even in the absence of additional data processing, yields relatively good results. This may be attributable to the ability of the embedding layer's nonlinearity to decode data non-monotonicities, which could be obscured when solely relying on covariance matrices. Further details regarding the ablation study can be found in the Appendix.

\paragraph{Efficiency.} Training of the neural network was executed in approximately 6 hours on an A-100 GPU with 81,920 MiB of graphical memory. Inference typically requires less a few seconds. In contrast many unsupervised approaches incur a significantly higher runtime. A time comparison is provided in Appendix \ref{AppendixTime}.

\section{Conclusion}
\label{Conclusion}
In this study, we introduced a novel neural network model for supervised graph structure learning, addressing the identifiability issue in observational data and modeling dependence relations through random Chebyshev polynomial dependencies. We introduced an observational-to-dependency processing approach, operating in both the Euclidean observational space and the \ac{spd} covariance manifold. Our model incorporates a novel bilinear attention mechanism and a permutation- and shape-agnostic architecture. We also establish theoretical properties confirming the model's robustness. Simulations demonstrate that BAM outperforms existing models in various scenarios.

BAM operates on transformed data's covariance matrices, enabling explicit processing of dependency information in the \ac{spd} matrix manifold, suggesting its potential applicability to other optimization challenges on the SPD manifold. Also, the observation-to-dependency processing model opens promising avenues for research, especially in applications requiring an understanding of variable dependencies.

\bibliography{BAM_references}

\newpage

\onecolumn

\noindent\hrule height4pt
\vskip .25in
\begin{center}
    \Large \bfseries Graph Structure Inference with BAM: \\Introducing the Bilinear Attention Mechanism \\(Appendix)
\end{center}
\noindent\hrule height1pt
\vskip .25in

\appendix
\section{Architectural Details}
\subsection{SPD Activation Function}\label{AppendixActivation}
To obtain an activation function for the \ac{spd} net, we leverage the following theorem, a direct consequence of \citePos{Guillot2015}{Theorem 4.11}, which is based on the work of \citet{schoenberg1942}:
\begin{theorem}
\label{t1}
Any continuous function from $C([-1,1],[-1,1])$ acting elementwise on a matrix preserves positive definiteness if it can be expressed as a series $f:[-1,1]\to [-1,1]$ with
\begin{gather*}
f(x)=\sum_{k=1}^{\infty}x^k w_k, \\
\quad \text{\rm{subject to}} \quad  \sum_{k=1}^{\infty}w_k\leq 1 \quad \text{\rm{with}} \quad w_k \geq 0,\quad  \forall k\in\mathbb{N}.
\end{gather*}
\end{theorem}
We employ this theorem to construct an activation function for the \ac{spd} neural network, using a relatively small maximal polynomial degree value of $N_{\text{max}}=3$. This function employs trainable weights $w_k$ and is applied after 'correlation normalization', i.e., the conversion of covariance matrices into correlation matrices. Trainable activation functions using low-degree Taylor polynomials were also proposed in \citep{Hoon2016} for general neural networks, not focusing on \ac{spd} data. Additionally, \citep{Apicella2020ASO} provides various types of trainable activation functions for Euclidean neural networks.

\subsection{Observational Attention} \label{AppendixObsAtt}
This layer accepts an input $\bs{X}\in\mathbb{R}^{M\times d\times C}$, from which it generates keys $\bs{K}=\bs{X}\bs{W}_K\in\mathbb{R}^{M\times d\times c}$, queries $\bs{Q}=\bs{X}\bs{W}_Q\in\mathbb{R}^{M\times d\times c}$, and values $\bs{V}=\bs{X}\bs{W}_V\in\mathbb{R}^{M\times d\times C}$, with $\bs{W}_K,\bs{W}_Q\in\mathbb{R}^{C\times c}$, and $\bs{W}_V\in\mathbb{R}^{C\times C}$.

For each $m=1,\dots, M$, keys and queries are combined in parallel along the inner axis, leading to
\begin{equation*}
\bs{K}\odot\bs{Q}:=\Nleft(\bs{K}_{m,\cdot,\cdot} \bs{Q}_{m,\cdot,\cdot}\Nright)_{m=1,\dots,M}\in\mathbb{R}^{M\times d\times d}.
\end{equation*}

This results in the attention weights
\begin{equation*}
\bs{A}=\sigma\Nleft(\frac{\bs{K}\odot\bs{Q}^{T(1,3,2)}}{\sqrt{c}}\Nright)\in\mathbb{R}^{M\times d\times d},
\end{equation*}
where $T(\operatorname{perm})$ denotes a permutation of the axes according to a permutation $\operatorname{perm}$ and $\sigma$ is the softmax operation along the last axis. For example, if $\text{perm}=[1,3,2]$, the operation would rearrange the second and third axes of the tensor while the first axis stays unchanged. Finally, for each $m$, the output is computed as 
\[\bs{H}=\bs{A}\odot \bs{V}\in\mathbb{R}^{M\times d\times C}.\]

\begin{figure}
    \centering
    \includegraphics[width=\textwidth]{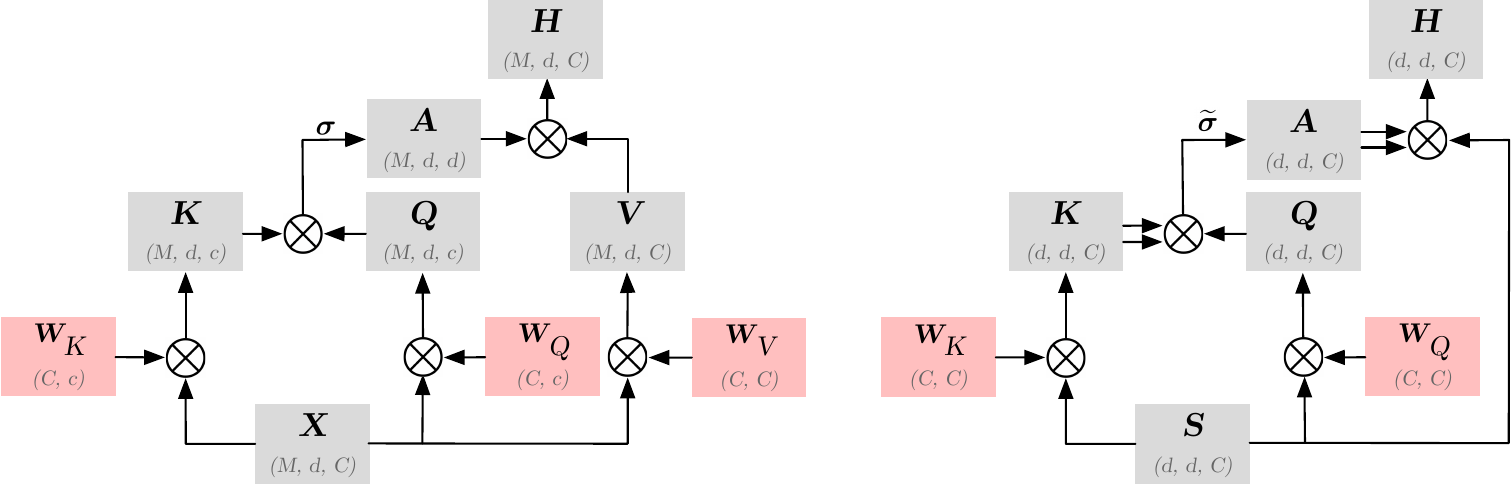}
    \caption{Left: Observational data self-attention layer across attributes. Gray denotes non-trainable tensors, and red represents trainable weights. The model-channel dimension, $c$, is chosen to be smaller than $C$ for efficiency. Matrix multiplication is performed after necessary transposition to match axis dimensions. Right: Bilinear self-attention layer. Gray indicates non-trainable tensors, and red denotes trainable weights. The double arrow signifies the use of the matrix as a bilinear operator.}
    \label{AttentionGraficObsBi_fig}
\end{figure}

\subsection{Residual Connections and Normalization.}\label{AppendixResidual}
Normalization is critical in attention networks \citep{Xiong2020}, with various methodologies available \citep{wu2018group, ba2016layer}. We adopt normalization and a residual connection \citep{he2016deep} on the input of all attention layers, expressed as $\bs{S}+\operatorname{Attention}(\operatorname{LayerNorm}(\bs{S}))$. Particularly when $M<d$, the residual connection can enhance the rank of covariance matrices, so full-rank representations can be attained even for under-determined problems. For learnable residual scaling, we utilize methods from \citet{touvron2021going, bachlechner2021rezero}.

Within the \ac{spd} manifold, we employ correlation normalization alongside residual connections. This approach preserves positive definiteness and corresponds intuitively to standard normalization in Euclidean space.

\subsection{Multiple Heads}\label{AppendixHeads}
In all attention layers, we employ multihead attention, a process that divides the input tensor along the channel axis into several smaller tensors. Each of these is then subjected to attention independently.

\newpage
\section{Theoretical Foundation for the Three-Class Edge Classification Problem}\label{Appendix3Class}

Building upon the foundational principles of Markov and faithfulness \citep{koller2009probabilistic}, we demonstrate that the three-class classification problem can be theoretically deduced from the distribution of the nodes by examining the following independence relations:
If there is an directed edge from node $X$ to $Y$ in the \ac{dag}, then $X$ and $Y$ are dependent given any set from the power set of the other nodes, i.e.:
\begin{equation*}
 X\rightarrow Y \implies X \notindependent Y \mid \bs{C} \quad \forall \bs{C}\in \mathcal{P}\Nleft(\mathcal{V}\setminus \{X, Y\}\Nright).
 \end{equation*}
 For an immorality between $X$ and $Y$, there exists a set of nodes $\bs{C}\in\mathcal{P}\Nleft(\mathcal{V}\setminus \{X, Y\}\Nright)$ within the power set of all other nodes such that $X$ and $Y$ are conditionally independent given this set (e.g., the set of all common ancestors of $X$ and $Y$, or the set of all parents of $X$ or $Y$), but dependent given all other nodes in the graph, i.e.:
  \begin{align*}
 &X\rightarrow Z \leftarrow Y, X \sNeg[-1mu]{\leftarrow}Y, X \sNeg[1mu]{\rightarrow} Y 
 \implies \exists \bs{C}\in \mathcal{P}\Nleft(\mathcal{V}\setminus \{X, Y\}\Nright): 
  X \independent Y \mid \bs{C}, X \notindependent Y \mid \mathcal{V}\setminus\{X,Y \}.
  \end{align*}
  If there is no edge between $X$ and $Y$, and if $X$ and $Y$ do not have a common child, then $X$ and $Y$ are conditionally independent given all other nodes, i.e.,
\begin{align*}
 &X \sNeg[-1mu]{\leftarrow}Y, X \sNeg[1mu]{\rightarrow} Y, \quad \nexists Z: X\rightarrow Z \leftarrow Y
 \quad\implies X \independent Y \mid \mathcal{V}\setminus\{X,Y \}.
\end{align*}

Note that testing for the no-edge class is cost-effective, as one only needs to test for a single set, rather than checking for any $Z \in \mathcal{V}\setminus\{X,Y\}$ if there is a v-structure $X\rightarrow Z \leftarrow Y$. 
Testing for a sepset to differentiate between the skeleton and moralized edge classes is more intricate. The neural network is tasked with learning an approximation for this distinction.

\newpage 
\section{Details of the CPDAG Estimation Model}\label{AppendixCPDAG}
\begin{figure}[H]
\centering
\includegraphics[height=0.83\textheight]{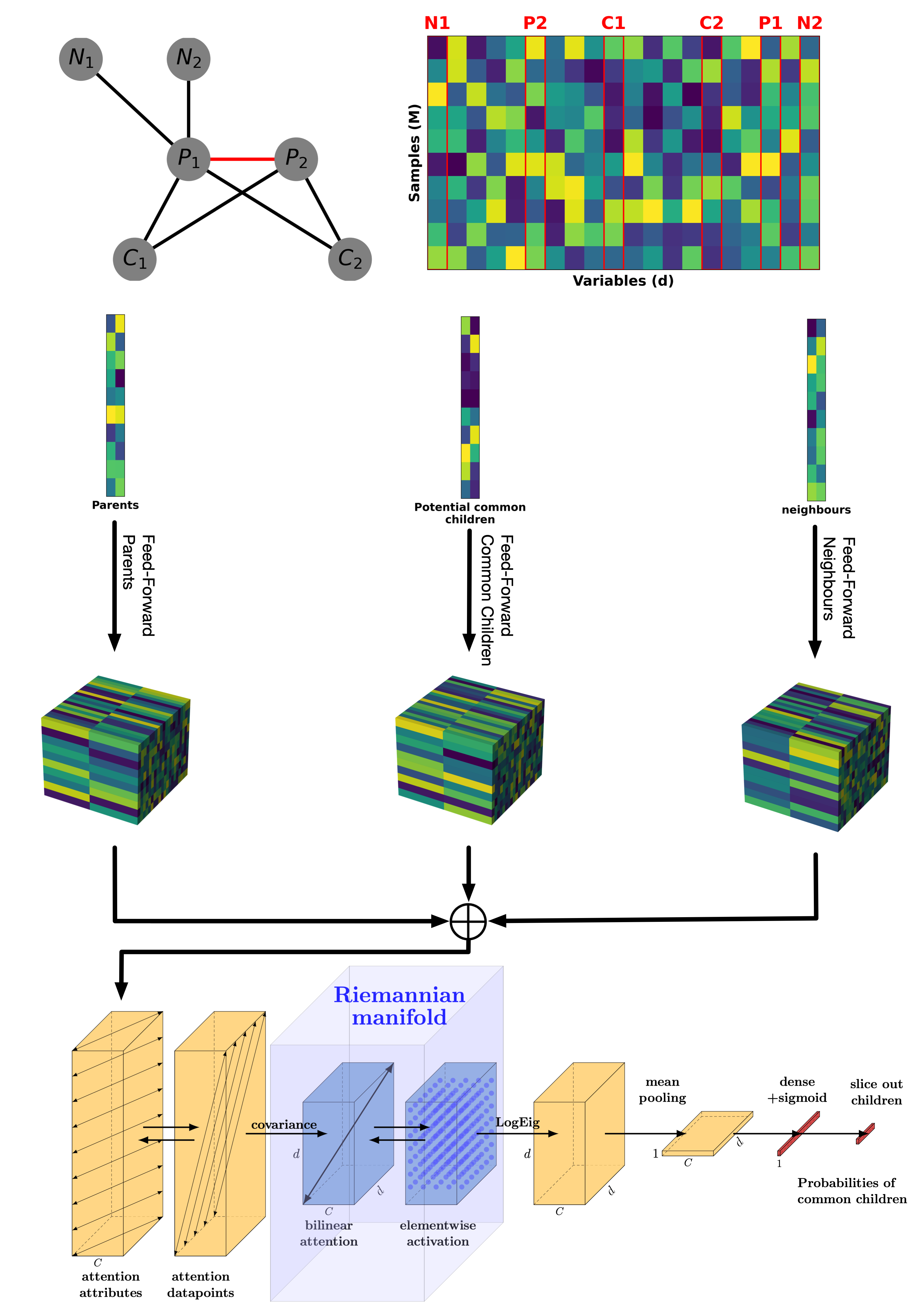}
\caption{CPDAG Estimation Architecture: In the graph, black edges represent undirected connections, whereas the red edge signifies an immorality. Columns in the data matrix corresponding to parent nodes, potential common children, and neighboring nodes are taken as input for the neural network and are processed through three distinct feed-forward networks. The embeddings are concatenated along the variable axis and processed through a network with data matrix attention, bilinear attention, and a LogEig layer. Mean pooling and sigmoid activation are applied to output probabilities for potential common children.
}
\label{figCPDAG}
\end{figure}

To estimate a CPDAG from the graph skeleton, along with the set of immoralities between pairs of nodes, we test each estimated immorality to determine which potential common child nodes are indeed common children. The parent nodes, denoted by $\text{pa}$, are the nodes between which an immorality was first estimated. Potential common child nodes, denoted by $\text{cc}$, are nodes that have an edge to both parent nodes. Neighbor nodes, denoted by $\text{ne}$, are nodes that have one edge to exactly one parent. 

We take the columns $\bs{x}_{\text{pa}}\in\mathbb{R}^{M\times 2}, \bs{x}_{\text{cc}}\in\mathbb{R}^{M\times |\text{cc}|}, \bs{x}_{\text{ne}}\in\mathbb{R}^{M\times |\text{ne}|}$ of the data matrix $\bs{X}$ corresponding to $\text{pa}, \text{cc}, \text{ne}$ as inputs for the neural network. Each of these submatrices undergoes a dimensionality expansion to $\bt{x}_{\text{pa}}\in\mathbb{R}^{M\times 2\times 1}, \bt{x}_{\text{cc}}\in\mathbb{R}^{M\times |\text{cc}|\times 1}, \bt{x}_{\text{ne}}\in\mathbb{R}^{M\times |\text{ne}|\times 1}$. Next, three separate feed-forward subnetworks $l_{\text{pa}}, l_{\text{cc}}, l_{\text{ne}}$ are applied to the three inputs $\bs{x}_{\text{pa}}, \bs{x}_{\text{cc}}, \bs{x}_{\text{ne}}$. Each of these layers has the same architecture: For an input $\bt{x}\in\mathbb{R}^{M\times d\times 1}$, weight matrices $\bs{W}_1\in\mathbb{R}^{1\times C}$ and $\bt{W}_1\in\mathbb{R}^{C\times C}$, together with a bias vector $\bs{b}_1\in\mathbb{R}^{C}$ are used to embed $\bt{x}$ to 
\[\bs{h}_1=\tanh\Nleft(\bt{x}\bs{W}_1+\bs{b}_1\Nright)\bt{W}_1\in\mathbb{R}^{M\times d\times C}\]
Then, two residual layers of the form 
\[\bs{h}_{i+1}=\bs{h}_i+\tanh\Nleft(\bs{h}_i\bs{W}_i+\bs{b}_i\Nright)\bt{W}_i\]
with $\bs{W}_i\in\mathbb{R}^{C\times C}$, $\bt{W}_i\in{R}^{C\times C}$, $\bs{b}_i\in\mathbb{R}^C$, $i=1, 2$ are applied to obtain representations $\bs{h}_{\text{pa}}\in\mathbb{R}^{M\times 2\times C},\bs{h}_{\text{cc}}\in\mathbb{R}^{M\times |\text{cc}|\times C}, \bs{h}_{\text{ne}}\in\mathbb{R}^{M\times |\text{ne}|\times C}$. For the addition of the bias terms, broadcasting is used, i.e., \[\bs{b}\in\mathbb{R}^{C} \mapsto \bt{b}\in \mathbb{R}^{M\times d\times C} \text{ with }\bt{b}_{m,l,c}=\bs{b}_c \quad  m=1,\dots, M,\text{ } l=1\dots, d,\text{ } c=1,\dots, C.\]

Now, the representations are concatenated along the dimension axis to obtain a $M\times (2+|\text{cc}|+|\text{ne}|)\times C$ tensor. We use $d=2+|\text{cc}|+|\text{ne}|$. We employ the same observation-to-dependency network as before, but instead of using softmax on the output of the LogEig-Layer, we use mean-pooling
\[\bs{X}\mapsto \left(\frac{1}{d}\bs{X}^{T(3,2,1)}\bs{1}_{d}\right)^T\]
to inflate one of the variable axes of the $\mathbb{R}^{d\times d\times C}$ output of LogEig to obtain a $\mathbb{R}^{d\times C}$ batch of $C$ vectors of dimension $d$. After applying a dense $C\times 1$ layer, we obtain a vector of length $d$.  Now, the entries corresponding to the potential common children can be sliced out and backpropagated for training.

The network architecture is shown in figure \ref{figCPDAG}.
\newpage

\section{Parameterization of the SEM}\label{AppendixSEM}
\subsection{Chebyshev Polynomials for Training}
For the Chebyshev polynomial, we utilize the following parameterization:
\begin{align*}
f_v(\bs{x}_{\text{pa}}, \epsilon_1,\epsilon_2) &= \sum_{w\in\text{pa}_{\mathcal{G}}\!(v)}\beta_w\sum_{n=1}^{r}\alpha_{n}T_n(x_w)+\epsilon_1 \\
&+ \alpha_{\text{m}}\left(\sum_{s,t\in\text{pa}(\mathcal{G}),s<t}\delta_{s,t} T_{\text{m}}(x_s,x_t)+\sum_{w\in\text{pa}(\mathcal{G})}T_{\text{m}}(x_w,\epsilon_2)\right) \quad\forall v \in \mathcal V,
\end{align*} where $r=5$ is the degree, and $T_n$ denotes the Chebyshev polynomials of the first kind (scaled for the input to have a maximum absolute value of $1$), and bivariate polynomials are given by:
\begin{equation}\label{EQChebMult}
T_{\text{m}}(x,y):=\frac{(x-\mu_x)(y-\mu_y)}{(1+\left|\mu_x\right|)(1+\left|\mu_y\right|)}
\end{equation} 
where $\text{m}$ in the index stands for "multidimensional".

Here, $\mu_x\sim U[-1,1]$, $\mu_y\sim U[-1,1]$, and the coefficients $\alpha_{1},\dots, \alpha_{r}, \alpha_{\text{m}}$ are calculated from $\alpha_{i}=\frac{\widetilde{\alpha}_{i}}{\sum_n \widetilde{\alpha}_{n}+\widetilde{\alpha}_{\text{m}}}$ for $i=1,\dots, r$, with $\widetilde{\alpha}_{i}=\frac{\gamma_{i}}{i!}$, $\gamma_{i}\sim U[-1,1]$, and $\widetilde{\alpha}_{\text{m}}\sim U[-1,1]$. $\beta_w=\frac{\widetilde{\beta}_w}{|\text{pa}|}$ with $\widetilde{\beta}_w\sim U[0.7,1.3]$. $\delta_{s,t}$ are random weights with $\delta_{s,t}=\frac{\widetilde{\delta}_{s,t}}{\sum_{s,t\in\text{pa}(\mathcal{G}),s<t} |\widetilde{\delta}_{s,t}|}$, $\widetilde{\delta}_{s,t}\sim U[-1,1]$

This parameterization is motivated by the observation that for a smooth function—where higher-order derivatives are not significantly larger than the lower-order ones—the coefficients of the Chebyshev approximation decrease in a factorial manner, as noted by \citet{xiang2020}. Rapidly decreasing Chebyshev coefficients were also observed by \citet{Trefethen2008}. This provides a rationale for training the neural network on 'typical' smooth functional dependencies. Furthermore, this suggests that using Chebyshev polynomials of degree $r=5$ is not a significant limitation, as the coefficients of higher orders are already negligibly small.

While multivariate Chebyshev polynomials constructed using Weyl-Groups were considered in \citep{Hofman1988,puschel2007}, we argue that our multivariate terms behave more nicely since they are bounded between $[-1,1]$, akin to univariate Chebychev polynomials. Their construction also intuitively incorporates multiplicative effects of two variables in a randomly shifted way.

In order to prevent the values from exploding and to properly account for the common domain of Chebyshev polynomials, we implement several measures. Firstly, we scale each input to the SEM by the maximum value within the batch. Secondly, we standardize all variables; we subtract the mean and divide by the standard deviation for each batch. This ensures the variables are both centered and scaled. To further improve stability and robustness of our model, we introduce thresholds for any absolute values exceeding $5$, thereby mitigating the potential impact of outliers.

\subsection{Gaussian Mixture Error Terms}
The additive error term $\epsilon_1$ follows a Gaussian mixture distribution. We randomly determine the number of components $L\sim U\{1,\dots, 5\}$ from a discrete uniform distribution. Each component has randomly assigned parameters for the means $\mu_l\sim U[-1,1]$, standard deviations $\sigma_l\sim U[0.05,1]$, and weights $\widetilde{w}_l\sim U[0.3,1]$, $w_l=\frac{\widetilde{w}_l}{\sum_l \widetilde{w}_l}$ such that $\epsilon_1\sim \sum_{l=1}^L w_l N(\mu_l,\sigma_l^2)$. The multiplicative error term $\epsilon_2$ is uniformly distributed, with $\epsilon_2\sim U[-1,1]$.

\subsection{Testing Dependencies}
To create testing data, we create synthetic data according to an SEM equipped with different dependency function, while the error term follows a Gaussian mixture distribution as before. We use the following dependencies for testing: Chebyshev, linear, sine, cosine, $x^2$, $x^3$, multidimensional multiplicative dependency. The Chebyshev-dependency used was the same as in the training procedure. We used the following testing dependency functions: $x$, $\sin(x)$, $\cos(x)$, $x^2$, $x^3$ as $g(x)$ in 

\[f_{\text{test}}(\bs{x}_{\text{pa}},\varepsilon)=\sum_{w\in \text{pa}(\mathcal{G})} \alpha_w g(x_w)+\epsilon .\]

For the multi-dimensional multiplicative test dependency, we used 
\[
f_v(\bs{x}_{\text{pa}}, \epsilon_1,\epsilon_2) = \alpha_{\text{m}}\left(\sum_{s,t\in\text{pa}(\mathcal{G}),s<t}\delta_{s,t} T_{\text{m}}(x_s,x_t)+\sum_{w\in\text{pa}(\mathcal{G})}T_{\text{m}}(x_w,\epsilon_2)\right) \quad\forall v \in \mathcal V,
\]
with 
\[
T_{\text{m}}(x,y):=\frac{(x-\mu_x)(y-\mu_y)}{(1+\left|\mu_x\right|)(1+\left|\mu_y\right|)}
\]
with $\mu_x\sim U[-1,1]$, $\mu_y\sim U[-1,1]$, $\delta_{s,t}=\frac{\widetilde{\delta}_{s,t}}{\sum_{s,t\in\text{pa}(\mathcal{G}),s<t} |\widetilde{\delta}_{s,t}|}$, $\widetilde{\delta}_{s,t}\sim U[-1,1]$.

\newpage

\section{Training Details}
\subsection{Model Hyperparameters}
The implementation was performed using TensorFlow \citep{tensorflow2015}. We employed the ADAM optimizer by \citet{kingma2015Adam}. For training, we used the hyperparameters stated in Table \ref{tabLargeModel}:

\begin{table}[ht]
  \centering
  \begin{minipage}{0.45\textwidth}
  \centering
\caption{Hyperparameters for the undirected graph estimation}
\label{tabLargeModel}
  \begin{tabular}{lc}
    Hyperparameter  & Value\\
    \hline\\
    \textbf{Layer Parameters}\\
    Number of channels $C$&$100$\\
    Number of inner channels $c$&$100$\\
    Maximal degree activation function &$3$\\
    Attention heads& 5\\
    \\
    \textbf{Number of layers}&\\
    Attention between attributes&10\\
    Attention between samples&10\\
    $C\times C$ dense observational layers&10\\
    bilinear attention + \ac{spd} activation & 10\\
    \\
    \textbf{Training Schedule}\\
    epochs&$1000$\\
    samples per epoch & $128$\\
    Initial learning rate& $0.0005$\\
    Learning rate decrease factor & $\left(\frac{1}{10}\right)^{1/500}$ \\
    Minibatchsize & 1\\
  \end{tabular}
  \end{minipage}\hfill
  \begin{minipage}{0.45\textwidth}
    \centering
    \caption{Hyperparameters of the CPDAG estimation model}
  \begin{tabular}{lc}
    Hyperparameter  & Value\\
    \hline\\
    \textbf{Layer Parameters}\\
    Number of channels $C$&$100$\\
    Number of inner channels $c$&$100$\\
    Maximal degree activation function &$3$\\
    Attention heads& 5\\
    \\
    \textbf{Number of layers}&\\
    Attention between attributes&10\\
    Attention between samples&10\\
    $C\times C$ dense observational layers&10\\
    bilinear attention + \ac{spd} activation & 10\\
    \\
    \textbf{Training Schedule}\\
    epochs&$1000$\\
    matrices per epoch & $1$\\
    Initial learning rate& $0.0005$\\
    Learning rate decrease factor & $\left(\frac{1}{10}\right)^{1/1000}$ \\
    Minibatchsize & 1\\
  \end{tabular}
\label{tabCPDAGModel}  
\end{minipage}
\end{table}

Additionally, we generated data with a random number of samples $M\sim U\{50,51,\dots, 1000\}$ and a random variable dimension $d\sim U\{10,11,\dots, 100\}$.

\paragraph{Ablation Study.}
In the ablation study, we evaluate the performance of the full model in comparison to models with reduced complexities. The full model is comprised of two attention between attributes layers, two attention between samples layers, two dense layers, and four bilinear layers equipped with SPD activation functions. This setup maintains parity between the number of attention layers operating on observational data and those focusing on covariance data, while also ensuring a comparable parameter count across different configurations. All models in the study utilize $C=c=100$ channels and are trained over $500$ epochs, with each epoch comprising $128$ data matrix / adjacency label pairs. Again, we generated data with a random number of samples $M\sim U\{50,51,\dots, 1000\}$ and a random variable dimension $d\sim U\{10,11,\dots, 100\}$.

\subsection{Loss Function for the Three-Class Edge Classification Problem.}
We employ the categorical cross-entropy loss function for classifying edges into one of three categories: \emph{no-edge}, \emph{skeleton edge}, and \emph{moralized edge}. Additionally, to enforce the condition that a moral edge between nodes $X$ and $Y$ should only be predicted if there is a potential common child $Z$ (i.e., $X - Z$ and $Z - Y$), we introduce a penalty term, $\mathcal{L}_\text{p}$.

The overall loss function is defined as $\mathcal{L}_{\text{b}}+\mathcal{L}_{\text{c}}+\mathcal{L}_{\text{p}}$. Here,  $\mathcal{L}_{\text{c}}$ is the categorical crossentropy of the three categories given by
 \[\mathcal{L}_{\text{c}}(\boldsymbol{A}, \widehat{\boldsymbol{A}}):= H(\boldsymbol{A}, \widehat{\boldsymbol{A}}) = - \sum_{i=2}^{d}\sum_{j=1}^{i-1}\sum_{c=1}^3 A_{i,j,c} \log(\widehat{A}_{i,j,c})\] denotes the categorical crossentropy of the three categories \emph{no-edge}, \emph{skeleton edge}, and \emph{moralized edge} with 
\[\bs{A}\in\mathbb{R}^{d\times d\times 3}\quad \text{with}\quad A_{i,j,c}=\begin{cases}
    1 & \text{if $(i,j)$ is in category $c$ in the ground-truth DAG}\\
    0   & \text{else}
\end{cases}\]
and $\widehat{A}_{i,j,c}$ denotes the estimation by the algorithm on it. $\bs{A}$ and $\widehat{\bs{A}}$ are symmetric along its first two axes, i.e., $A_{i,j,c}=A_{j,i,c},\quad i,j=1,\dots, d, \quad c=1,\dots, 3$.

$\mathcal{L}_{\text{b}}$ denotes the binary loss of no-edge vs. any edge present (present edges $=$ skeleton edges $\cup$ moralized edges):
\[\mathcal{L}_{\text{b}}(\boldsymbol{A}^{(b)}, \widehat{\boldsymbol{A}}^{(b)}):= H(\boldsymbol{A}^{(b)}, \widehat{\boldsymbol{A}}^{(b)}) = - \sum_{i=2}^{d}\sum_{j=1}^i \left[ A_{i,j}^{(b)} \log(\widehat{A}_{i,j}^{(b)}) + (1 - A_{i,j}^{(b)}) \log(1 - \widehat{A}_{i,j}^{(b)}) \right]\]
with
\[\bs{A}^{(b)}\in \mathbb{R}^{d\times d}\quad \text{with}\quad A_{i,j}^{(b)}=\begin{cases}
    1 & \text{if an edge is estimated between $i$ and $j$}\\
    0   & \text{else}
\end{cases}\] being the adjacency matrix of no-edge vs. (direct edge $\cup$ moralized edge).

The penalty term, $\mathcal{L}_{\text{p}}$ is defined as:
\[\mathcal{L}_{\text{p}}(\widehat{\boldsymbol{A}}):=\max\left(\widehat{\boldsymbol{A}}_3-[\widehat{\boldsymbol{A}}_2\widehat{\boldsymbol{A}}_2]^{0.5},0\right).\] 
$\widehat{\boldsymbol{A}}_1, \widehat{\boldsymbol{A}}_2, \widehat{\boldsymbol{A}}_3\in\mathbb{R}^{d\times d}$ are estimates of $\boldsymbol{A}_{\cdot,\cdot,1}$, $\boldsymbol{A}_{\cdot,\cdot,2}$, and $\boldsymbol{A}_{\cdot,\cdot,3}$ by the algorithm respectively. This term penalizes the prediction of a moralized edge in the absence of potential common child edges. The square root operation is applied element-wise.
\newpage

\section{Further Experiments}
\subsection{Error Bars}
In our experiments, each algorithm was evaluated on five distinct trials, each involving a unique data matrix and corresponding ground-truth graph. The bars in the figures represent the mean performance values across these trials, while the error bars indicate the standard deviation. It is important to note that the observed variability, manifested as relatively large error bars, is predominantly due to the random sampling of graph degrees, which has a substantial influence on estimation accuracy. Despite this inherent variability, the comparison across algorithms remains valid, as each algorithm is tested for the same graphs. Therefore, the magnitude of the error bars should not be interpreted as undermining the reliability of our findings.

\subsection{Additional Results on Undirected Graph Estimation}

\begin{figure}[ht!]
\centering
\includegraphics[width=\textwidth]{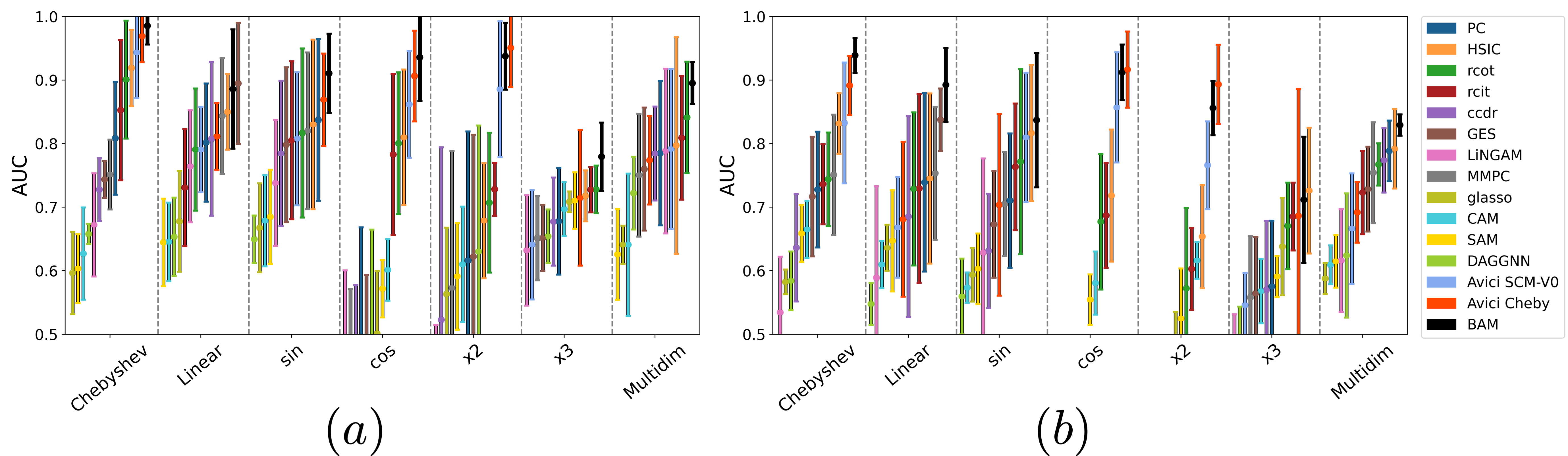}
\caption{AUC values for undirected graph estimation in low-dimensional regimes: (a) $d=10$, $M=200$ (b)$d=20$, $M=500$.}
\label{figAddUndirected}
\end{figure}

\begin{figure}[ht!]
\centering
\includegraphics[width=\textwidth]{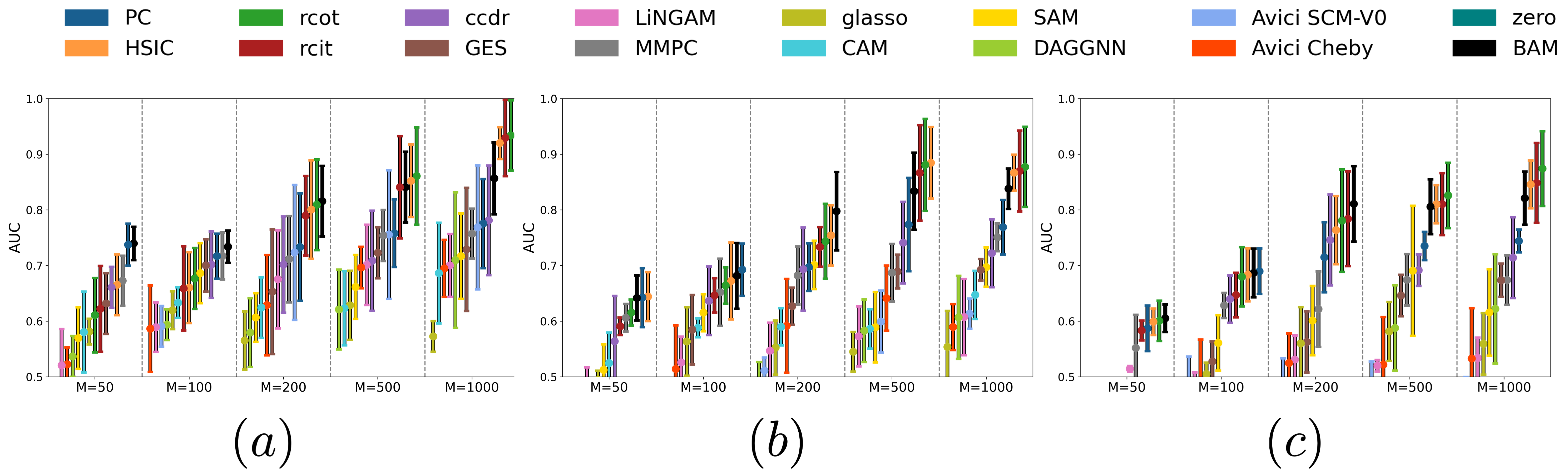}
\caption{AUC values for undirected graph estimation for random MLP dependency: (a): $d=20$, (b): $d=50$, (c): $d=100$}
\label{figAddUndirectedMLP}
\end{figure}

Figures \ref{figAddUndirected} and \ref{figAddUndirectedMLP} provide supplemental data on the task of undirected graph estimation. Figures \ref{figAddUndirected} showcases performance in low-dimensional settings characterized by $d=10$, $M=200$ and $d=20$, $M=500$. In this scenario, our method (BAM) also outperforms competing graph inference algorithms. To further assess its capability to recognize multidimensional dependencies, we extended our tests to cases where the dependency function within the SEM is modeled via a randomly initialized \ac{mlp} with a random number of layers $\sim \mathcal U\{1,\dots,5\}$, a random number of hidden layers $\sim \mathcal U\{4,64\}$, and $\operatorname{relu}$ or $\operatorname{tanh}$ activation with probability $0.5$ each. Despite these complexities, our algorithm maintained state-of-the-art performance, delivering AUC scores competitive to the top-performing existing methods such as depicted in Figure \ref{figAddUndirectedMLP}.

\subsection{Additional Results on CPDAG Estimation}\label{AppendixDirectedResults}

\begin{figure}[ht!]
\centering
\includegraphics[width=\textwidth]{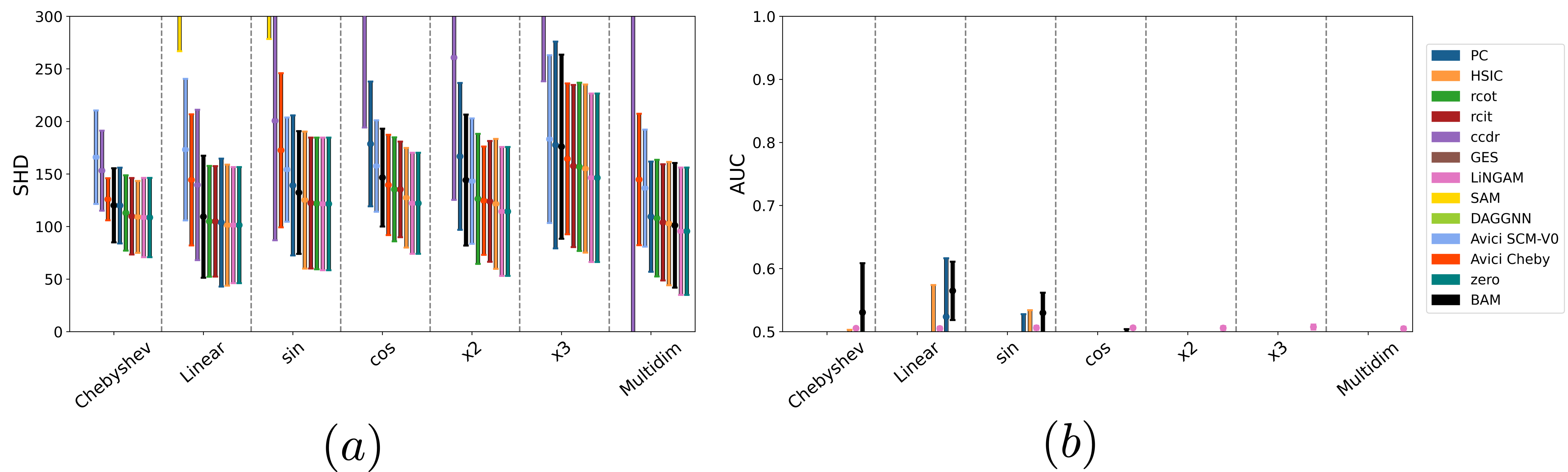}
\caption{(a) SHD values for the high-dimensional CPDAG estimation $d=100$, $M=50$. (b) AUC values for the high-dimensional CPDAG estimation $d=100$, $M=50$.}
\label{figAddDirected}
\end{figure}

CPDAG estimation in high-dimensional settings presents significant challenges. In the specific case of $d=100$ and $M=50$, none of the algorithms we evaluated could outperform a zero-graph (i.e., a graph with no edges) baseline in terms of Structural Hamming Distance (SHD). The results, depicted in Figure \ref{figAddDirected} (a), substantiate this observation and suggest that CPDAG estimation remains a difficult problem under these conditions, at least with our chosen graph density setup. Given the complexities encountered in high-dimensional contexts, our analysis primarily emphasizes the evaluation of AUC, as illustrated in Figure \ref{figAddDirected}(b). In this evaluation, only the BAM and PC algorithms demonstrated AUC values exceeding $0.5$ in certain instances. However, the AUC metrics remain low. This underscores the utility of undirected graph methods for high-dimensional ($d>M$) problems for such problems, as directed approaches may not only be inefficient but also risk yielding misleading interpretations. 

\subsection{Time Comparison}\label{AppendixTime}
\begin{figure}[h!]
\centering
\includegraphics[width=0.95\textwidth]{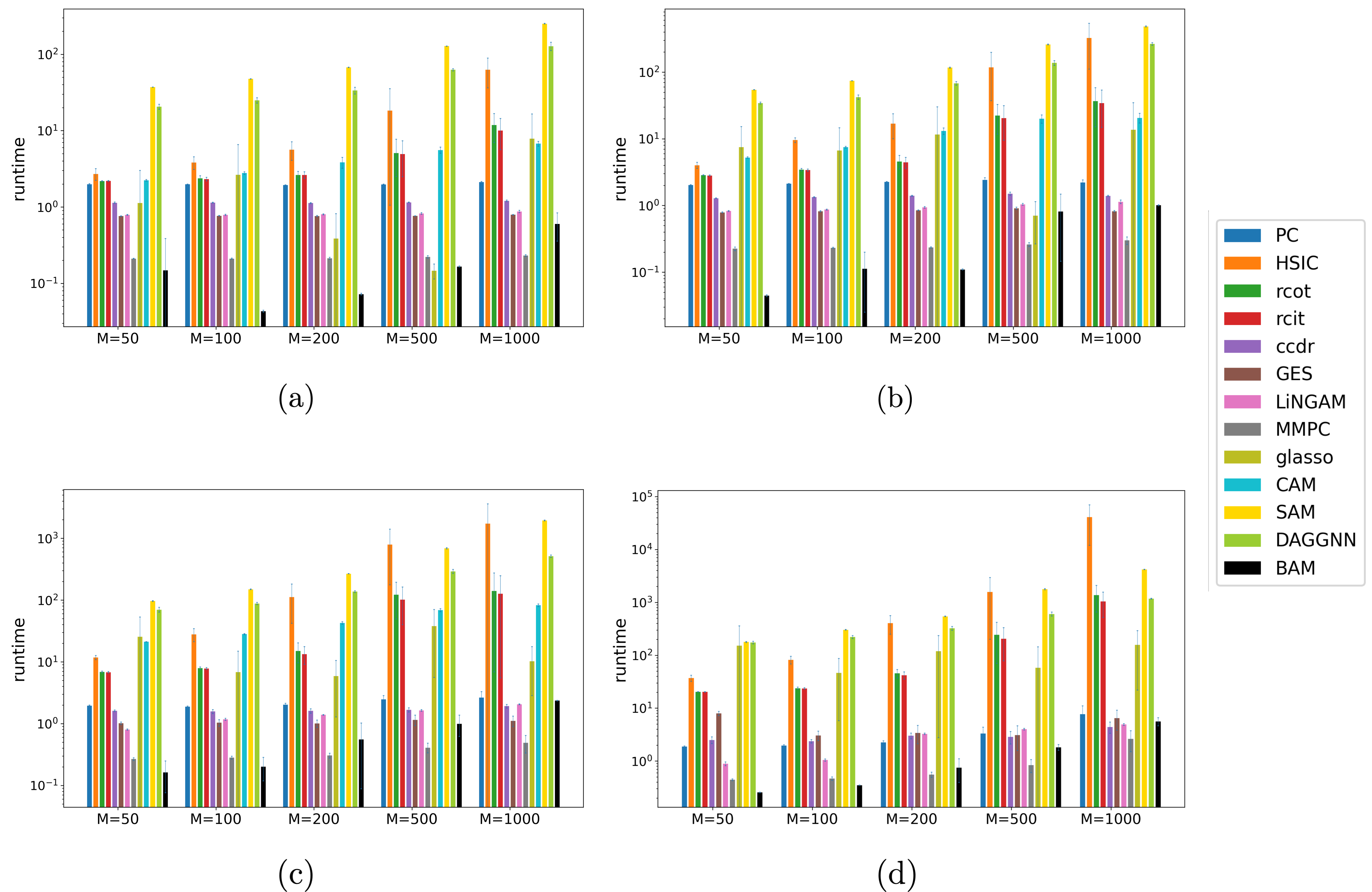}
\caption{Algorithm runtime for (a) $d=10$, (b) $d=20$, (c) $d=50$, (d) $d=100$ in seconds per $M\times d$ data matrix inference.
}
\label{figRuntimes}
\end{figure}

Figure \ref{figRuntimes} depicts the average runtimes per evaluation step, accompanied by their corresponding standard deviations. We compared BAM with various unsupervised methods, noting that the evaluation times for other supervised approaches, such as Avici, are comparable to those observed for BAM. Specifically, the results are presented in the form of mean $\pm$ one standard deviation. The x-axis enumerates various sample sizes, denoted as $M$, while both mean and standard deviation were computed based on $5$ independent inference tests for each configuration with a fixed sample size $M$ and graph dimension $d$.

These empirical observations substantiate the computational efficiency of supervised approaches in the inference phase.

\newpage
\section{Interpretation}\label{AppendixIntuition}
\subsection{Shape-Agnostic Architecture and the Role of Attention Layers}
When employing a shape-agnostic architecture for matrices $\in\mathbb{R}^{M\times d}$, it is crucial to ensure that all elements within the $M\times d$ matrix can interact and influence one another. Consider a scenario where one axis of the matrix is expanded to the shape $\mathbb{R}^{M\times d\times 1}$, followed by dense layers with $1\times C$ and $C\times C$ weights. In this configuration, the dense layers carry out element-wise operations on the $M\times d$ elements, processing them in isolation from each other. This is because each hidden representation is essentially a linear combination of $C$ matrices of shape $M\times d$ prior to the application of an element-wise activation function.

This limitation is addressed by incorporating attention layers into the architecture. These layers adaptively compute non-trainable $M\times M$ and $d\times d$ attention matrices based on trainable $C\times C$ weights. This approach allows for a permutation- and shape-agnostic architecture, as the same set of trainable weights can be employed for any $M\times d$ matrix, while still enabling the matrix entries to influence each other. In this way, the attention mechanism becomes an essential component of our model. Although we also experimented with Network-in-Network methods \citep{lin2013network}, we found that the attention mechanism offers a more stable, efficient, and straightforward computation of non-trainable $M\times M$ and $d\times d$ attention matrices, using only trainable $C\times C$ weights.

Our proposed bilinear attention mechanism is, to our knowledge, the first SPD layer to enable shape-agnostic computations. It uses trainable $C\times C$ weights to calculate non-trainable $d\times d$ attention matrices, allowing for adaptive weighting across different $d\times d$ SPD matrix sizes. This flexibility makes it a unique and essential component of our architecture. Additionally, this construction ensures the desired permutation invariance among the variables. Our approach essentially learns matrix operations that should be applicable to any input matrix with arbitrary $M\times d$ input. 

\subsection{Attention scores in the BAM layer}
Consider the setting as in Figure 3 (right) and the computation of the output $\bs{H}$ by $\bs{A}\otimes \bs{S}$, where $\bs{A}\in \spaceSPDC$ are the attention scores and $\bs{S}\in \spaceSPDC$ are the input matrices into the BAM layer. Since $\bs{A}\otimes \bs{S}$ is processed parallel across the channels, we consider for simplicity the output of a single channel here and assume $\bs{A}\in\spaceSPD$ and $\bs{S}\in\spaceSPD$ to be quadratic, positive definite $d\times d$ matrices.

While traditional self-attention computes scores to assess the importance of one data point to another, our bilinear attention mechanism extends this by exploring the interdependence of variable pairs. Specifically, for an output pair $(i,j)$, its associated output value is determined not merely by a direct scalar relationship but by the bilinear form: $\sum_{k,l} A_{i,k}S_{k,l}A_{l,j}$. Thus, instead of a singular focus on the relation "How does $j$ affect $i$?", quantified in the score matrix $\bs{A}$ in classical attention, the score matrix in bilinear attention shows the interaction strengths of pair sets $\{(i,k)\mid k=1,\dots,d\}$ and $\{(l,j)\mid l=1,\dots, d\}$. The "receptive field" adopts a cross-form within the scores $\bs{A}$ instead of being $A_{i,j}$ only, in the sense that relevant scores for the output at position $(i,j)$ are not limited to $A_{i,j}$ but $\{\bs{A}_{i,\cdot}\cup \bs{A}_{\cdot, j}\}$.

\subsection{Keys and Queries}
Continuing with the single-channel assumption due to parallel channel processing, consider the quadratic form $\mathcal{S}^{d\times d}\times\mathcal{S}^{d\times d}\mapsto \mathcal{S}^{d\times d}$, $(\bs{K} ,\bs{Q} )\mapsto \bs{K}^T\bs{Q} \bs{K} $ of the key-query interaction. The $(i,j)$-th entry of $\bs{K}^T \bs{Q} \bs{K}$ is $\bs{K}^T_i \bs{Q} \bs{K}_j$ for the columns $\bs{K}_1,\dots , \bs{K}_d$ of $\bs{K}$, which are often referred to as keys. Using the eigendecomposition of $\bs{Q}=\bs{U}^T \bs{D} \bs{U} $ one obtains for the $(i,j)$-th entry the bilinear form $(\bs{U} \bs{K}_i)^T \bs{D} (\bs{U} \bs{K}_j)$. Note that $\bs{U} \bs{K}_i$ is a similarity measure between $\bs{K}$ and $\bs{Q}$ analogous to standard attention. So, for bilinear attention, similarity scores are calculated between the keys $\bs{K}$ and the eigenvectors of the queries $\bs{Q}$. Afterwards, the $\bs{D} ^{\frac{1}{2}}$-weighted bilinear-form $(\bs{D}^{\frac{1}{2}}\bs{U} \bs{K}_i)^T(\bs{D} ^{\frac{1}{2}}\bs{U} \bs{K}_j)$ is used to create covariance matrices by combining the similarity scores between $\bs{U}$ and $\bs{K}$. Hence, in bilinear attention, the similarity scores are functions of both the $i$-th and $j$-th keys as well as all queries. This is consistent with the attention-score behavior, where the interaction strengths of all pair sets $\{(i,k)\mid k=1,\dots,d\}$ and $\{(l,j)\mid l=1,\dots, d\}$ collectively influence the output.

This is in contrast to standard attention, which uses the untransformed dot product $(\bs{k} , \bs{q} )\in\mathbb{R}^d\times \mathbb{R}^d\mapsto \bs{k}^T \bs{q}\in\mathbb{R} $ for the columns of key and query matrices $\bs{K}$, $\bs{Q}$. 

\newpage
\section{Limitations}
The model effectively captures smooth dependence relations using Chebyshev polynomials. Although this approach excels across various types of dependencies, it might have limitations for data structures that deviate significantly from the generated synthetic data. However, adapting the synthetic data generation to accommodate these structures is straightforward.

As for the Log-Eig layer, it performs efficiently within a moderate dimensional range but may face computational challenges when scaling to higher dimensions. Training our model with parameters similar to those used in this study demands substantial memory resources; in our experiments, around 80 GB of GPU memory was required. 

The attention mechanism, while effective, can be costly for high-dimensional (both, in $M$ and $d$) inputs. This can be ameliorated with local attention, although this approach may introduce its own set of challenges. While the model is effective for its intended applications, its architecture allows for easy extensions. For instance, a separate embedding layer could be added for both observational and interventional data to make use of interventional data.

An end-to-end approach for CPDAG estimation might offer further benefits. The current model loses directional information in the covariance computation, making an end-to-end approach for CPDAG estimation unfeasible with the existing architecture. However, a simple extension could involve using two separate embeddings for each variable, one for the variable as parent, one for the variable as child, effectively doubling the dimensionality to $2d$, to potentially facilitate directional inference.  Lastly, like other neural network-based approaches, there is a potential risk of overfitting (here on Chebyshev polynomial dependencies), necessitating hyperparameter tuning.

\end{document}

%% file: acronyms.tex
\begin{acronym}
    \acro{mjp}[MJP]{Markov jump process}
    \acroplural{mjp}[MJPs]{Markov jump processes}
    \acroindefinite{mjp}{an}{a}
    \acro{dag}[DAG]{directed acyclic graph}
     \acro{msa}[MSA]{multi-sequence alignment}
      \acroindefinite{msa}{an}{a}
\acro{glasso}[Glasso]{graphical lasso}
     \acro{spd}[SPD]{symmetric positive semi-definite}
      \acroindefinite{spd}{an}{a}
      \acro{sem}[SEM]{structural equation model}
      \acroindefinite{sem}{an}{a}
       \acro{nlp}[NLP]{natural language processing}
      \acroindefinite{nlp}{an}{a}
      \acro{er}[ER]{Erdős–Rényi}
      \acroindefinite{er}{an}{an}
      \acro{mlp}[MLP]{multilayer perceptron}
    \acroindefinite{mlp}{an}{a}
    \acro{mrf}[MRF]{Markov random field}
    \acroindefinite{mrf}{an}{a}
\end{acronym}

%% file: BAM.bbl
\begin{thebibliography}{75}
\providecommand{\natexlab}[1]{#1}
\providecommand{\url}[1]{\texttt{#1}}
\expandafter\ifx\csname urlstyle\endcsname\relax
  \providecommand{\doi}[1]{doi: #1}\else
  \providecommand{\doi}{doi: \begingroup \urlstyle{rm}\Url}\fi

\bibitem[Abadi et~al.(2015)Abadi, Agarwal, Barham, and et~al.]{tensorflow2015}
Mart\'{i}n Abadi, Ashish Agarwal, Paul Barham, and et~al.
\newblock {TensorFlow}: Large-scale machine learning on heterogeneous systems,
  2015.
\newblock URL \url{https://www.tensorflow.org/}.
\newblock Software available from tensorflow.org.

\bibitem[Absil et~al.(2008)Absil, Mahony, and Sepulchre]{absil2008optimization}
P-A Absil, Robert Mahony, and Rodolphe Sepulchre.
\newblock \emph{Optimization algorithms on matrix manifolds}.
\newblock Princeton University Press, 2008.

\bibitem[Apicella et~al.(2021)Apicella, Donnarumma, Isgr{\`o}, and
  Prevete]{Apicella2020ASO}
Andrea Apicella, Francesco Donnarumma, Francesco Isgr{\`o}, and Roberto
  Prevete.
\newblock {A survey on modern trainable activation functions}.
\newblock \emph{{Neural Networks}}, 138:\penalty0 14--32, 2021.

\bibitem[Aragam and Zhou(2015)]{aragam2015concave}
Bryon Aragam and Qing Zhou.
\newblock Concave penalized estimation of sparse gaussian bayesian networks.
\newblock \emph{The Journal of Machine Learning Research}, 16\penalty0
  (1):\penalty0 2273--2328, 2015.

\bibitem[Ba et~al.(2016)Ba, Kiros, and Hinton]{ba2016layer}
Jimmy~Lei Ba, Jamie~Ryan Kiros, and Geoffrey~E Hinton.
\newblock {Layer Normalization}.
\newblock \emph{{arXiv:1607.06450}}, 2016.

\bibitem[Bachlechner et~al.(2021)Bachlechner, Majumder, Mao, Cottrell, and
  McAuley]{bachlechner2021rezero}
Thomas Bachlechner, Bodhisattwa~Prasad Majumder, Henry Mao, Gary Cottrell, and
  Julian McAuley.
\newblock {ReZero is All You Need: Fast Convergence at Large Depth}.
\newblock In \emph{{UAI}}, pages 1352--1361, 2021.

\bibitem[Bahdanau et~al.(2015)Bahdanau, Cho, and Bengio]{Bahdanau2015}
Dzmitry Bahdanau, Kyunghyun Cho, and Yoshua Bengio.
\newblock {Neural Machine Translation by Jointly Learning to Align and
  Translate}.
\newblock In \emph{{ICLR}}, 2015.

\bibitem[Barfuss et~al.(2016)Barfuss, Massara, Di~Matteo, and
  Aste]{barfuss2016parsimonious}
Wolfram Barfuss, Guido~Previde Massara, T.~Di~Matteo, and Tomaso Aste.
\newblock {Parsimonious modeling with information filtering networks}.
\newblock \emph{{Phys. Rev. E}}, 94, 2016.

\bibitem[Bhatia(2009)]{bhatia2009}
Rajendra Bhatia.
\newblock {Positive Definite Matrices}.
\newblock In \emph{{Positive Definite Matrices}}. {Princeton University Press},
  2009.

\bibitem[Brouillard et~al.(2020)Brouillard, Lachapelle, Lacoste,
  Lacoste-Julien, and Drouin]{Brouillard2020}
Philippe Brouillard, S\'{e}bastien Lachapelle, Alexandre Lacoste, Simon
  Lacoste-Julien, and Alexandre Drouin.
\newblock {Differentiable Causal Discovery from Interventional Data}.
\newblock In \emph{{NeurIPS}}, volume~33, pages 21865--21877, 2020.

\bibitem[B\"{u}hlmann et~al.(2014)B\"{u}hlmann, Kalisch, and
  Meier]{Buehlmann2014annual}
Peter B\"{u}hlmann, Markus Kalisch, and Lukas Meier.
\newblock {High-Dimensional Statistics with a View Toward Applications in
  Biology}.
\newblock \emph{{Annu. Rev. Stat. Appl.}}, 1\penalty0 (1):\penalty0 255--278,
  2014.

\bibitem[B{\"u}hlmann et~al.(2014)B{\"u}hlmann, Peters, and
  Ernest]{buhlmann2014cam}
Peter B{\"u}hlmann, Jonas Peters, and Jan Ernest.
\newblock Cam: Causal additive models, high-dimensional order search and
  penalized regression.
\newblock \emph{The Annals of Statistics}, 2014.

\bibitem[Chickering(2002)]{chickering2002optimal}
David~Maxwell Chickering.
\newblock Optimal structure identification with greedy search.
\newblock \emph{Journal of machine learning research}, 3\penalty0
  (Nov):\penalty0 507--554, 2002.

\bibitem[Chung et~al.(2016)Chung, Lee, and Park]{Hoon2016}
Hoon Chung, Sung~Joo Lee, and Jeon~Gue Park.
\newblock {Deep Neural Network Using Trainable Activation Functions}.
\newblock In \emph{{IJCNN}}, pages 348--352, 2016.

\bibitem[Dai et~al.(2023)Dai, Ding, Jiang, Han, and Zhang]{Dai2023}
Haoyue Dai, Rui Ding, Yuanyuan Jiang, Shi Han, and Dongmei Zhang.
\newblock {ML4C: Seeing Causality Through Latent Vicinity}.
\newblock In \emph{{Proceedings of the 2023 SIAM International Conference on
  Data Mining (SDM)}}, pages 226--234, 2023.

\bibitem[Drton and Maathuis(2017)]{drton2017structure}
Mathias Drton and Marloes~H. Maathuis.
\newblock Structure learning in graphical modeling.
\newblock \emph{{Annual Review of Statistics and Its Application}}, 4\penalty0
  (1):\penalty0 365--393, 2017.

\bibitem[Friedman et~al.(2008)Friedman, Hastie, and Tibshirani]{Friedman2008}
Jerome Friedman, Trevor Hastie, and Robert Tibshirani.
\newblock {Sparse inverse covariance estimation with the graphical lasso}.
\newblock \emph{{Biostatistics}}, 9\penalty0 (3):\penalty0 432--441, 2008.

\bibitem[Gerstenberg et~al.(2021)Gerstenberg, Goodman, Lagnado, and
  Tenenbaum]{Gerstenberg2020}
Tobias Gerstenberg, Noah~D Goodman, David~A Lagnado, and Joshua~B Tenenbaum.
\newblock {A Counterfactual Simulation Model of Causal Judgments for Physical
  Events}.
\newblock \emph{{Psychol. Rev.}}, 128\penalty0 (5):\penalty0 936, 2021.

\bibitem[Glymour et~al.(2019)Glymour, Zhang, and Spirtes]{Glymour2019}
Clark Glymour, Kun Zhang, and Peter Spirtes.
\newblock Review of causal discovery methods based on graphical models.
\newblock \emph{Frontiers in Genetics}, 10:\penalty0 524, 2019.

\bibitem[Goudet et~al.(2018)Goudet, Kalainathan, Caillou, Guyon,
  et~al.]{goudet2018learning}
Olivier Goudet, Diviyan Kalainathan, Philippe Caillou, Isabelle Guyon, et~al.
\newblock {Learning Functional Causal Models with Generative Neural Networks}.
\newblock In \emph{{Explainable and Interpretable Models in Computer Vision and
  Machine Learning}}, pages 39--80, 2018.

\bibitem[Guillot and Rajaratnam(2015)]{Guillot2015}
D.~Guillot and B.~Rajaratnam.
\newblock {Functions Preserving Positive Definiteness for Sparse Matrices}.
\newblock \emph{{Proc. Am. Math. Soc.}}, 367:\penalty0 627--649, 2015.

\bibitem[Hauser and B{\"u}hlmann(2012)]{hauser2012characterization}
Alain Hauser and Peter B{\"u}hlmann.
\newblock Characterization and greedy learning of interventional markov
  equivalence classes of directed acyclic graphs.
\newblock \emph{The Journal of Machine Learning Research}, 13\penalty0
  (1):\penalty0 2409--2464, 2012.

\bibitem[He and Garcia(2009)]{he2009Learning}
Haibo He and Edwardo~A. Garcia.
\newblock Learning from imbalanced data.
\newblock \emph{IEEE Transactions on Knowledge and Data Engineering},
  21\penalty0 (9):\penalty0 1263--1284, 2009.
\newblock \doi{10.1109/TKDE.2008.239}.

\bibitem[He et~al.(2016)He, Zhang, Ren, and Sun]{he2016deep}
Kaiming He, Xiangyu Zhang, Shaoqing Ren, and Jian Sun.
\newblock {Deep Residual Learning for Image Recognition}.
\newblock In \emph{{CVPR}}, pages 770--778, 2016.

\bibitem[Ho et~al.(2019)Ho, Kalchbrenner, Weissenborn, and Salimans]{Ho2019}
Jonathan Ho, Nal Kalchbrenner, Dirk Weissenborn, and Tim Salimans.
\newblock {Axial Attention in Multidimensional Transformers}.
\newblock \emph{{arXiv:1912.12180}}, 2019.

\bibitem[Hoffman and Withers(1988)]{Hofman1988}
Michael~E. Hoffman and William~Douglas Withers.
\newblock Generalized chebyshev polynomials associated with affine weyl groups.
\newblock \emph{{Transactions of the American Mathematical Society}},
  308\penalty0 (1):\penalty0 91--104, 1988.

\bibitem[Hsieh et~al.(2014)Hsieh, Sustik, Dhillon, and Ravikumar]{cho2014quic}
Cho-Jui Hsieh, M{{\'a}}ty{{\'a}}s~A. Sustik, Inderjit~S. Dhillon, and Pradeep
  Ravikumar.
\newblock {QUIC: Quadratic Approximation for Sparse Inverse Covariance
  Estimation}.
\newblock \emph{{J. Mach. Learn. Res.}}, 15\penalty0 (83):\penalty0 2911--2947,
  2014.

\bibitem[Huang and Gool(2017)]{huang2017}
Zhiwu Huang and Luc~Van Gool.
\newblock {A Riemannian Network for SPD Matrix Learning}.
\newblock In \emph{{AAAI}}, page 2036–2042, 2017.

\bibitem[Hyttinen et~al.(2013)Hyttinen, Eberhardt, Hoyer, and
  Jarvisalo]{hyttinen2013}
A.~Hyttinen, F.~Eberhardt, O.~Hoyer, and M.~Jarvisalo.
\newblock Discovering cyclic causal models with latent variables: a general
  sat-based procedure.
\newblock In \emph{Proceedings of the 29th Conference on Uncertainty in
  Artificial Intelligence}, 2013.

\bibitem[Jones et~al.(2012)Jones, Buchan, Cozzetto, and Pontil]{Jones2012}
David~T Jones, Daniel W~A Buchan, Domenico Cozzetto, and Massimiliano Pontil.
\newblock {PSICOV: precise structural contact prediction using sparse inverse
  covariance estimation on large multiple sequence alignments}.
\newblock \emph{{Bioinformatics}}, 28\penalty0 (2):\penalty0 184--190, 2012.

\bibitem[Kalainathan et~al.(2020)Kalainathan, Goudet, and
  Dutta]{kalainathan2020causal}
Diviyan Kalainathan, Olivier Goudet, and Ritik Dutta.
\newblock Causal discovery toolbox: Uncovering causal relationships in python.
\newblock \emph{The Journal of Machine Learning Research}, 21\penalty0
  (1):\penalty0 1406--1410, 2020.

\bibitem[Kalainathan et~al.(2022)Kalainathan, Goudet, Guyon, Lopez-Paz, and
  Sebag]{Kalainathan2022}
Diviyan Kalainathan, Olivier Goudet, Isabelle Guyon, David Lopez-Paz, and
  Mich{\`e}le Sebag.
\newblock {Structural Agnostic Modeling: Adversarial Learning of Causal
  Graphs}.
\newblock \emph{{J. Mach. Learn. Res.}}, 23\penalty0 (219):\penalty0 1--62,
  2022.

\bibitem[Ke et~al.(2022)Ke, Chiappa, Wang, Bornschein, Goyal, Rey, Botvinick,
  Weber, et~al.]{ke2022learning}
Nan~Rosemary Ke, Silvia Chiappa, Jane~X Wang, Jorg Bornschein, Anirudh Goyal,
  Melanie Rey, Matthew Botvinick, Theophane Weber, et~al.
\newblock {Learning to induce causal structure}.
\newblock In \emph{{ICML: Workshop on Spurious Correlations, Invariance and
  Stability}}, 2022.

\bibitem[Khan et~al.(2022)Khan, Naseer, Hayat, Zamir, Khan, and Shah]{Khan2022}
Salman Khan, Muzammal Naseer, Munawar Hayat, Syed~Waqas Zamir, Fahad~Shahbaz
  Khan, and Mubarak Shah.
\newblock {Transformers in Vision: A Survey}.
\newblock \emph{{ACM Comput. Surv.}}, 54\penalty0 (10s), 2022.

\bibitem[Kingma and Ba(2015)]{kingma2015Adam}
Diederik~P. Kingma and Jimmy Ba.
\newblock Adam: {A} method for stochastic optimization.
\newblock In Yoshua Bengio and Yann LeCun, editors, \emph{3rd International
  Conference on Learning Representations, {ICLR} 2015, San Diego, CA, USA, May
  7-9, 2015, Conference Track Proceedings}, 2015.

\bibitem[Koller and Friedman(2009)]{koller2009probabilistic}
Daphne Koller and Nir Friedman.
\newblock \emph{{Probabilistic Graphical Models: Principles and Techniques }}.
\newblock MIT press, 2009.

\bibitem[Kossen et~al.(2021)Kossen, Band, Lyle, Gomez, Rainforth, and
  Gal]{kossen2021selfattention}
Jannik Kossen, Neil Band, Clare Lyle, Aidan~N Gomez, Thomas Rainforth, and
  Yarin Gal.
\newblock {Self-Attention Between Datapoints: Going Beyond Individual
  Input-Output Pairs in Deep Learning}.
\newblock In \emph{{NeurIPS}}, 2021.

\bibitem[Lauritzen and Spiegelhalter(1988)]{LauritzenLocal}
S.~L. Lauritzen and D.~J. Spiegelhalter.
\newblock Local computations with probabilities on graphical structures and
  their application to expert systems.
\newblock \emph{Journal of the Royal Statistical Society. Series B
  (Methodological)}, 50\penalty0 (2):\penalty0 157--224, 1988.
\newblock ISSN 00359246.

\bibitem[Li et~al.(2020)Li, Xiao, and Tian]{li2020supervised}
Hebi Li, Qi~Xiao, and Jin Tian.
\newblock {Supervised Whole DAG Causal Discovery}.
\newblock \emph{{arXiv:2006.04697}}, 2020.

\bibitem[Li et~al.(2019)Li, Hu, Zhang, Yu, and Zhang]{Li2019}
Yang Li, Jun Hu, Chengxin Zhang, Dong-Jun Yu, and Yang Zhang.
\newblock {ResPRE: high-accuracy protein contact prediction by coupling
  precision matrix with deep residual neural networks}.
\newblock \emph{{Bioinformatics}}, 35\penalty0 (22):\penalty0 4647--4655, 2019.

\bibitem[Lin et~al.(2013)Lin, Chen, and Yan]{lin2013network}
Min Lin, Qiang Chen, and Shuicheng Yan.
\newblock Network in network.
\newblock \emph{arXiv preprint arXiv:1312.4400}, 2013.

\bibitem[Lopez-Paz et~al.(2015{\natexlab{a}})Lopez-Paz, Muandet, and
  Recht]{lopez2015randomized}
David Lopez-Paz, Krikamol Muandet, and Benjamin Recht.
\newblock The randomized causation coefficient.
\newblock \emph{{JMLR}}, 16:\penalty0 2901--2907, 2015{\natexlab{a}}.

\bibitem[Lopez-Paz et~al.(2015{\natexlab{b}})Lopez-Paz, Muandet, Sch{\"o}lkopf,
  and Tolstikhin]{lopez2015towards}
David Lopez-Paz, Krikamol Muandet, Bernhard Sch{\"o}lkopf, and Iliya
  Tolstikhin.
\newblock Towards a learning theory of cause-effect inference.
\newblock In \emph{{ICML}}, pages 1452--1461, 2015{\natexlab{b}}.

\bibitem[Lopez-Paz et~al.(2017)Lopez-Paz, Nishihara, Chintala, Sch{\"o}lkopf,
  and Bottou]{Lopez2017}
David Lopez-Paz, Robert Nishihara, Soumith Chintala, Bernhard Sch{\"o}lkopf,
  and Leon Bottou.
\newblock Discovering causal signals in images.
\newblock In \emph{Proceedings of the IEEE Conference on Computer Vision and
  Pattern Recognition (CVPR)}, July 2017.

\bibitem[Lorch et~al.(2022)Lorch, Sussex, Rothfuss, Krause, and
  Sch\"{o}lkopf]{Lorch2022}
Lars Lorch, Scott Sussex, Jonas Rothfuss, Andreas Krause, and Bernhard
  Sch\"{o}lkopf.
\newblock Amortized inference for causal structure learning.
\newblock In S.~Koyejo, S.~Mohamed, A.~Agarwal, D.~Belgrave, K.~Cho, and A.~Oh,
  editors, \emph{Advances in Neural Information Processing Systems}, volume~35,
  pages 13104--13118. Curran Associates, Inc., 2022.

\bibitem[Ma et~al.(2022)Ma, Ding, Dai, Jiang, Wang, Han, and Zhang]{Ma2022}
Pingchuan Ma, Rui Ding, Haoyue Dai, Yuanyuan Jiang, Shuai Wang, Shi Han, and
  Dongmei Zhang.
\newblock {ML4S: Learning Causal Skeleton from Vicinal Graphs}.
\newblock In \emph{{SIGKDD}}, page 1213–1223, 2022.

\bibitem[Meek(1995)]{Meek1995}
Christopher Meek.
\newblock {Causal Inference and Causal Explanation with Background Knowledge}.
\newblock In \emph{{UAI}}, pages 403--410, 1995.

\bibitem[Nowack et~al.(2020)Nowack, Runge, Eyring, and Haigh]{nowack2020causal}
Peer Nowack, Jakob Runge, Veronika Eyring, and Joanna~D Haigh.
\newblock Causal networks for climate model evaluation and constrained
  projections.
\newblock \emph{Nature communications}, 11\penalty0 (1):\penalty0 1415, 2020.

\bibitem[Pedregosa et~al.(2011)Pedregosa, Varoquaux, Gramfort, Michel, Thirion,
  Grisel, Blondel, Prettenhofer, Weiss, Dubourg, Vanderplas, Passos,
  Cournapeau, Brucher, Perrot, and Duchesnay]{scikit-learn}
F.~Pedregosa, G.~Varoquaux, A.~Gramfort, V.~Michel, B.~Thirion, O.~Grisel,
  M.~Blondel, P.~Prettenhofer, R.~Weiss, V.~Dubourg, J.~Vanderplas, A.~Passos,
  D.~Cournapeau, M.~Brucher, M.~Perrot, and E.~Duchesnay.
\newblock Scikit-learn: Machine learning in {P}ython.
\newblock \emph{Journal of Machine Learning Research}, 12\penalty0
  (85):\penalty0 2825--2830, 2011.

\bibitem[Pennec et~al.(2006)Pennec, Fillard, and Ayache]{pennec2006riemannian}
Xavier Pennec, Pierre Fillard, and Nicholas Ayache.
\newblock {A Riemannian Framework for Tensor Computing}.
\newblock \emph{{Int. J. Comput. Vis.}}, 66\penalty0 (1):\penalty0 41--66,
  2006.

\bibitem[Peters et~al.(2017)Peters, Janzing, and
  Sch{\"o}lkopf]{peters2017elements}
Jonas Peters, Dominik Janzing, and Bernhard Sch{\"o}lkopf.
\newblock \emph{{Elements of Causal Inference}}.
\newblock {MIT Press}, Cambridge, Massachusetts, 2017.

\bibitem[Puschel and Rotteler(2007)]{puschel2007}
Markus Puschel and Martin Rotteler.
\newblock Algebraic signal processing theory: 2-d spatial hexagonal lattice.
\newblock \emph{{IEEE Transactions on Image Processing}}, 16\penalty0
  (6):\penalty0 1506--1521, 2007.

\bibitem[Rao et~al.(2021)Rao, Liu, Verkuil, Meier, Canny, Abbeel, Sercu, and
  Rives]{Rao2021}
Roshan~M Rao, Jason Liu, Robert Verkuil, Joshua Meier, John Canny, Pieter
  Abbeel, Tom Sercu, and Alexander Rives.
\newblock {MSA Transformer}.
\newblock In \emph{{ICML}}, volume 139, pages 8844--8856. PMLR, 2021.

\bibitem[Reynolds(2009)]{reynolds2009gaussian}
Douglas~A Reynolds.
\newblock Gaussian mixture models.
\newblock \emph{Encyclopedia of biometrics}, 741\penalty0 (659-663), 2009.

\bibitem[Schoenberg(1942)]{schoenberg1942}
I.~J. Schoenberg.
\newblock Positive definite functions on spheres.
\newblock \emph{Duke Mathematical Journal}, 9:\penalty0 96--108, 1942.
\newblock MR0005922 (3,232c).

\bibitem[Shah and Peters(2020)]{shah2020}
R.~D. Shah and J.~Peters.
\newblock The hardness of conditional independence testing and the generalised
  covariance measure.
\newblock \emph{The Annals of Statistics}, 48\penalty0 (3), 2020.

\bibitem[Shalom et~al.(2022)Shalom, Treister, and Yavneh]{shalom2022pista}
Gal Shalom, Eran Treister, and Irad Yavneh.
\newblock pista: preconditioned iterative soft thresholding algorithm for
  graphical lasso.
\newblock \emph{arXiv:2205.10027}, 2022.

\bibitem[Shimizu et~al.(2006)Shimizu, Hoyer, Hyv{\"a}rinen, Kerminen, and
  Jordan]{shimizu2006linear}
Shohei Shimizu, Patrik~O Hoyer, Aapo Hyv{\"a}rinen, Antti Kerminen, and Michael
  Jordan.
\newblock A linear non-gaussian acyclic model for causal discovery.
\newblock \emph{Journal of Machine Learning Research}, 7\penalty0 (10), 2006.

\bibitem[Song et~al.(2019)Song, Shi, Xiao, Duan, Xu, Zhang, and Tang]{song2019}
Weiping Song, Chence Shi, Zhiping Xiao, Zhijian Duan, Yewen Xu, Ming Zhang, and
  Jian Tang.
\newblock {AutoInt: Automatic Feature Interaction Learning via Self-Attentive
  Neural Networks}.
\newblock In \emph{{CIKM}}, pages 1161--1170, 2019.

\bibitem[Spirtes et~al.(2000{\natexlab{a}})Spirtes, Glymour, Scheines,
  Kauffman, Aimale, and Wimberly]{spirtes2000constructing}
Pater Spirtes, Clark Glymour, Richard Scheines, Stuart Kauffman, Valerio
  Aimale, and Frank Wimberly.
\newblock Constructing {B}ayesian network models of gene expression networks
  from microarray data.
\newblock In \emph{Proceedings of the Atlantic Symposium on Computational
  Biology}. Carnegie Mellon University, 2000{\natexlab{a}}.

\bibitem[Spirtes et~al.(2000{\natexlab{b}})Spirtes, Glymour, Scheines, and
  Heckerman]{spirtes2000causation}
Peter Spirtes, Clark~N Glymour, Richard Scheines, and David Heckerman.
\newblock \emph{{Causation, Prediction, and Search}}.
\newblock {MIT Press}, 2000{\natexlab{b}}.

\bibitem[Strobl et~al.(2019)Strobl, Zhang, and
  Visweswaran]{strobl2019approximate}
Eric~V Strobl, Kun Zhang, and Shyam Visweswaran.
\newblock Approximate kernel-based conditional independence tests for fast
  non-parametric causal discovery.
\newblock \emph{Journal of Causal Inference}, 7\penalty0 (1):\penalty0
  20180017, 2019.

\bibitem[Touvron et~al.(2021)Touvron, Cord, Sablayrolles, Synnaeve, and
  J{\'e}gou]{touvron2021going}
Hugo Touvron, Matthieu Cord, Alexandre Sablayrolles, Gabriel Synnaeve, and
  Herv{\'e} J{\'e}gou.
\newblock {Going deeper with Image Transformers}.
\newblock In \emph{{ICCV}}, pages 32--42, 2021.

\bibitem[Trefethen(2008)]{Trefethen2008}
Lloyd~N. Trefethen.
\newblock {Is Gauss Quadrature Better than Clenshaw--Curtis?}
\newblock \emph{{SIAM Review}}, 50\penalty0 (1):\penalty0 67--87, 2008.

\bibitem[Tsamardinos et~al.(2006)Tsamardinos, Brown, and
  Aliferis]{tsamardinos2006max}
Ioannis Tsamardinos, Laura~E Brown, and Constantin~F Aliferis.
\newblock The max-min hill-climbing bayesian network structure learning
  algorithm.
\newblock \emph{Machine learning}, 65:\penalty0 31--78, 2006.

\bibitem[Vaswani et~al.(2017)Vaswani, Shazeer, Parmar, Uszkoreit, Jones, Gomez,
  Kaiser, and Polosukhin]{Vaswani2017}
Ashish Vaswani, Noam Shazeer, Niki Parmar, Jakob Uszkoreit, Llion Jones,
  Aidan~N Gomez, \L~ukasz Kaiser, and Illia Polosukhin.
\newblock {Attention is All you Need}.
\newblock In \emph{{NeurIPS}}, volume~30, 2017.

\bibitem[Vowels et~al.(2022)Vowels, Camgoz, and Bowden]{Vowels2022}
Matthew~J. Vowels, Necati~Cihan Camgoz, and Richard Bowden.
\newblock {D'Ya Like DAGs? A Survey on Structure Learning and Causal
  Discovery}.
\newblock \emph{{ACM Comput. Surv.}}, 55\penalty0 (4):\penalty0 82:1--82:36,
  nov 2022.
\newblock ISSN 0360-0300.

\bibitem[Wang et~al.(2022)Wang, Wu, Xu, Hu, and Kittler]{Wang2022}
Rui Wang, Xiao-Jun Wu, Tianyang Xu, Cong Hu, and Josef Kittler.
\newblock {Deep Metric Learning on the SPD Manifold for Image Set
  Classification}.
\newblock \emph{{IEEE Trans. Circuits Syst. Video. Technol.}}, 2022.

\bibitem[Wright(1921)]{Wright1921CorrelationAndCausation}
Sewall Wright.
\newblock {Correlation and causation}.
\newblock \emph{{J. Agric. Res.}}, 20\penalty0 (7):\penalty0 557--585, 1921.

\bibitem[Wu and He(2018)]{wu2018group}
Yuxin Wu and Kaiming He.
\newblock {Group Normalization}.
\newblock In \emph{{ECCV}}, pages 3--19, 2018.

\bibitem[Xiang and Liu(2020)]{xiang2020}
Shuhuang Xiang and Guidong Liu.
\newblock Optimal decay rates on the asymptotics of orthogonal polynomial
  expansions for functions of limited regularities.
\newblock \emph{{Numerische Mathematik}}, 145\penalty0 (1):\penalty0 117--148,
  2020.

\bibitem[Xiong et~al.(2020)Xiong, Yang, He, Zheng, Zheng, Xing, Zhang, Lan,
  et~al.]{Xiong2020}
Ruibin Xiong, Yunchang Yang, Di~He, Kai Zheng, Shuxin Zheng, Chen Xing,
  Huishuai Zhang, Yanyan Lan, et~al.
\newblock {On Layer Normalization in the Transformer Architecture}.
\newblock In \emph{{ICML}}, volume 119, pages 10524--10533, 2020.

\bibitem[Yu et~al.(2019)Yu, Chen, Gao, and Yu]{yu2019dag}
Yue Yu, Jie Chen, Tian Gao, and Mo~Yu.
\newblock {DAG-GNN: DAG structure learning with graph neural networks}.
\newblock In \emph{{ICML}}, pages 7154--7163, 2019.

\bibitem[Zhang et~al.(2012)Zhang, Peters, Janzing, and
  Sch{\"o}lkopf]{zhang2012kernel}
Kun Zhang, Jonas Peters, Dominik Janzing, and Bernhard Sch{\"o}lkopf.
\newblock Kernel-based conditional independence test and application in causal
  discovery.
\newblock \emph{arXiv preprint arXiv:1202.3775}, 2012.

\bibitem[Zheng et~al.(2018)Zheng, Aragam, Ravikumar, and Xing]{zheng2018}
Xun Zheng, Bryon Aragam, Pradeep~K Ravikumar, and Eric~P Xing.
\newblock {DAGs with NO TEARS: Continuous Optimization for Structure Learning}.
\newblock In \emph{{NeurIPS}}, volume~31, 2018.

\end{thebibliography}
